\pdfoutput=1

\documentclass[11pt]{article}

\usepackage{ACL2023} 

\usepackage{times}
\usepackage{latexsym}
\usepackage{graphicx}
\usepackage{amsmath}
\usepackage{bbold}
\usepackage{multirow}
\usepackage{pifont}
\usepackage{float}
\usepackage{lipsum} 
\usepackage{placeins}
\usepackage{booktabs}
\usepackage{calrsfs}
\usepackage{todonotes}
\usepackage{caption}
\usepackage{subcaption}
\usepackage{cleveref}

\newcommand*{\affaddr}[1]{#1} 
\newcommand*{\affmark}[1][*]{\textsuperscript{#1}}
\newcommand*{\email}[1]{\texttt{#1}}
\newcommand\tab[1][0.5cm]{\hspace*{#1}}

\usepackage[T1]{fontenc}

\usepackage[utf8]{inputenc}

\usepackage{microtype}

\usepackage{inconsolata}

\title{Tackling Ambiguity with Images: \\Improved Multimodal Machine Translation and Contrastive Evaluation}

\author{%
Matthieu Futeral\affmark[1,2]~~~Cordelia Schmid\affmark[1,2]~~~Ivan Laptev\affmark[1,2]\\
~~~\textbf{Benoît Sagot}\affmark[1]~~~\textbf{Rachel Bawden}\affmark[1] \\
\affaddr{\affmark[1]Inria Paris}\\
\affaddr{\affmark[2]Département d’informatique de l’ENS, CNRS, PSL Research University}\\
\email{firstname.lastname@inria.fr}
}

\begin{document}
\maketitle
\begin{abstract}
One of the major challenges of machine translation (MT) is ambiguity, which can in some cases be resolved by accompanying context such as images. However, recent work in multimodal MT (MMT) has shown that obtaining improvements from images is challenging, limited not only by the difficulty of building effective cross-modal representations, but also by the lack of specific evaluation and training data. 
We present a new MMT approach based on a strong text-only MT model, which uses neural adapters, a novel guided self-attention mechanism and which is jointly trained on both visually-conditioned masking and MMT.
We also introduce CoMMuTE, a Contrastive Multilingual Multimodal Translation Evaluation set of ambiguous sentences and their possible translations, accompanied by disambiguating images corresponding to each translation.
Our approach obtains competitive results compared to strong text-only models on standard English$\rightarrow$French, English$\rightarrow$German and English$\rightarrow$Czech benchmarks and outperforms baselines and state-of-the-art MMT systems by a large margin on our contrastive test set. Our code\footnote{\href{https://github.com/MatthieuFP/VGAMT}{https://github.com/MatthieuFP/VGAMT}} and CoMMuTE\footnote{\href{https://github.com/MatthieuFP/CoMMuTE}{https://github.com/MatthieuFP/CoMMuTE}} are freely available.
\end{abstract}

\section{Introduction}
{}

\begin{figure}[!t]
    \centering
    \includegraphics[width=\linewidth]{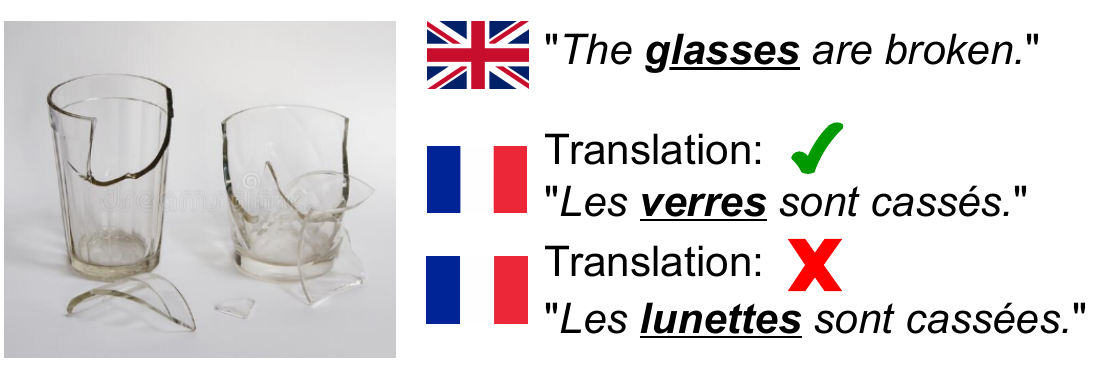}
    \caption{Visual context resolving the ambiguity of English word \textit{glasses} for English-to-French translation.
    }
    \label{fig:glasses_broken} \vspace{-1mm}
\end{figure}

Multimodal machine translation (MMT) typically refers to the use of additional non-textual data in text-based machine translation (MT). Here, we focus on the case where source texts are accompanied by images, the idea being to exploit visual data to improve the translation of ambiguous sentences.
For example, in Figure~\ref{fig:glasses_broken}, the English word \textit{glasses} can either be translated as French \textit{verres} `drinking vessels' or \textit{lunettes} `spectacles', an ambiguity which is resolved using the image. 

A main research direction of MMT has been how to best exploit image representations and combine the image and text modalities \citep{GRAPH-MMT, VTLM, doubly-calixto-2017, li2022valhalla}. It has typically been difficult to surpass strong text-only baselines, the image modality often being ignored \citep{good-for-misconceived-reasons-2021}. A major issue holding back progress is that most current state-of-the-art MMT models \citep{GRAPH-MMT, desmond-2017-imagination, good-for-misconceived-reasons-2021, li2022valhalla} are trained solely on the $\sim$30k examples of the Multi30k dataset \citep{elliott-2016-multi30k}, comprising image captions and their translations. This causes two issues: (i)~the models do not exploit the large amount of text-only data available 
and therefore perform poorly in comparison to state-of-the-art text-only MT systems, and (ii)~we show that very few examples require images to be correctly translated, which means that the datasets are ill-adapted to evaluating the use of the image modality.

In this article, we aim to overcome these problems by proposing (i)~a new MMT approach that is able to exploit (text-only) monolingual and parallel data as well as (multimodal) captioning data, and that reaches a good balance between maintaining high MT quality and effectively exploiting images, and (ii)~a test set, CoMMuTE, containing contrastive evaluation pairs, where images provide the necessary context to disambiguate between multiple meanings of the same source sentence.

Our suggested model is inspired by work on adapting frozen language models (LMs) to multimodal inputs \citep{Sung2022VLADAPTERPT,yang2022frozenbilm, eichenberg2021magma,pfeiffer2021xgqa}; we propose to adapt a strong MT model to multimodal inputs with lightweight modules \citep{houlsby2019parameter} to exploit the large amount of textual data it was trained on. We also propose to better exploit the image by introducing guided self-attention and by combining the standard MMT objective with a visually-conditioned masked language modelling (VMLM) objective \citep{li2019visualbert, lu2019vilbert, vlbert}. Our model obtains competitive results compared to strong text-only baselines on standard En$\rightarrow$\{Fr,De,Cs\} MMT benchmarks \citep{elliott-2016-multi30k,elliott-EtAl:2017:WMT,barrault2018findings} and outperforms them and state-of-the-art MMT models on our lexically ambiguous contrastive test set.\footnote{CoMMuTE initially contained 50 lexically ambiguous sentences in a previous version of the paper. Results are now computed on the updated version of CoMMuTE comprising 155 ambiguous sentences. Conclusions remain the same.}

\section{Related Work}

\paragraph{Multimodal MT data.} 
The reference dataset to train and evaluate MMT models is Multi30k \citep{elliott-2016-multi30k}. However, recent work has shown that most MMT systems trained and evaluated on it do not effectively exploit the image information; \citet{elliott-2018-adversarial} showed that replacing the ground truth image with a random one does not lead to the drop in performance that would be expected, while \citet{good-for-misconceived-reasons-2021} argued that the observed gain in performance was due to a regularisation effect. It is also notoriously difficult to beat text-only baselines on this benchmark \citep{barrault2018findings}. This may be due to (i)~some subsets of Multi30k having been translated independently from the images \citep{elliott-2016-multi30k} and (ii)~most of the time, the source text being sufficient in theory to produce a perfect translation (i.e.~the image is not necessary; see Section~\ref{data} for our own analysis).

Based on this, alternative test sets and evaluation methods have been proposed. \citet{caglayan-etal-2019-probing} proposed to probe the use of images in MMT models, while \citet{vision-matters} proposed another training corpus and evaluation benchmark to evaluate MMT systems, but their work is only based on gender ambiguity and requires specific training data to train MMT models. \citet{lala-specia-2018-multimodal} released a lexically ambiguous MMT evaluation dataset to evaluate models ability to disambiguate source sentences, but we found that text context is generally sufficient to translate the evaluation dataset correctly.

\paragraph{Contrastive MT datasets.}
Another means of evaluating (and the one we adopt here) is to target specific phenomena through the use of contrastive test sets. They involve evaluating models based on their ability to rank pairs of translations, where one is correct and the other incorrect.
They have been used for the evaluation of different linguistic phenomena, including grammaticality \citep{sennrich-2017-grammatical}, 
multi-sense word disambiguation \citep{rios-gonzales-etal-2017-improving, mucow}, pronoun translation \citep{contrapro,bawden-etal-2018-evaluating,voita-etal-2019-good} and lexical coherence/consistency \citep{bawden-etal-2018-evaluating,voita-etal-2019-good}. \citet{bawden-etal-2018-evaluating} introduced the idea of conditioning which of the translations is correct depending on linguistic context, and we adopt the same strategy here with our CoMMuTE dataset, composed of 
lexically ambiguous sentences whose translations are determined by the visual context.

\begin{figure*}[!th]
    \centering\small
    \includegraphics[width=0.95\linewidth]{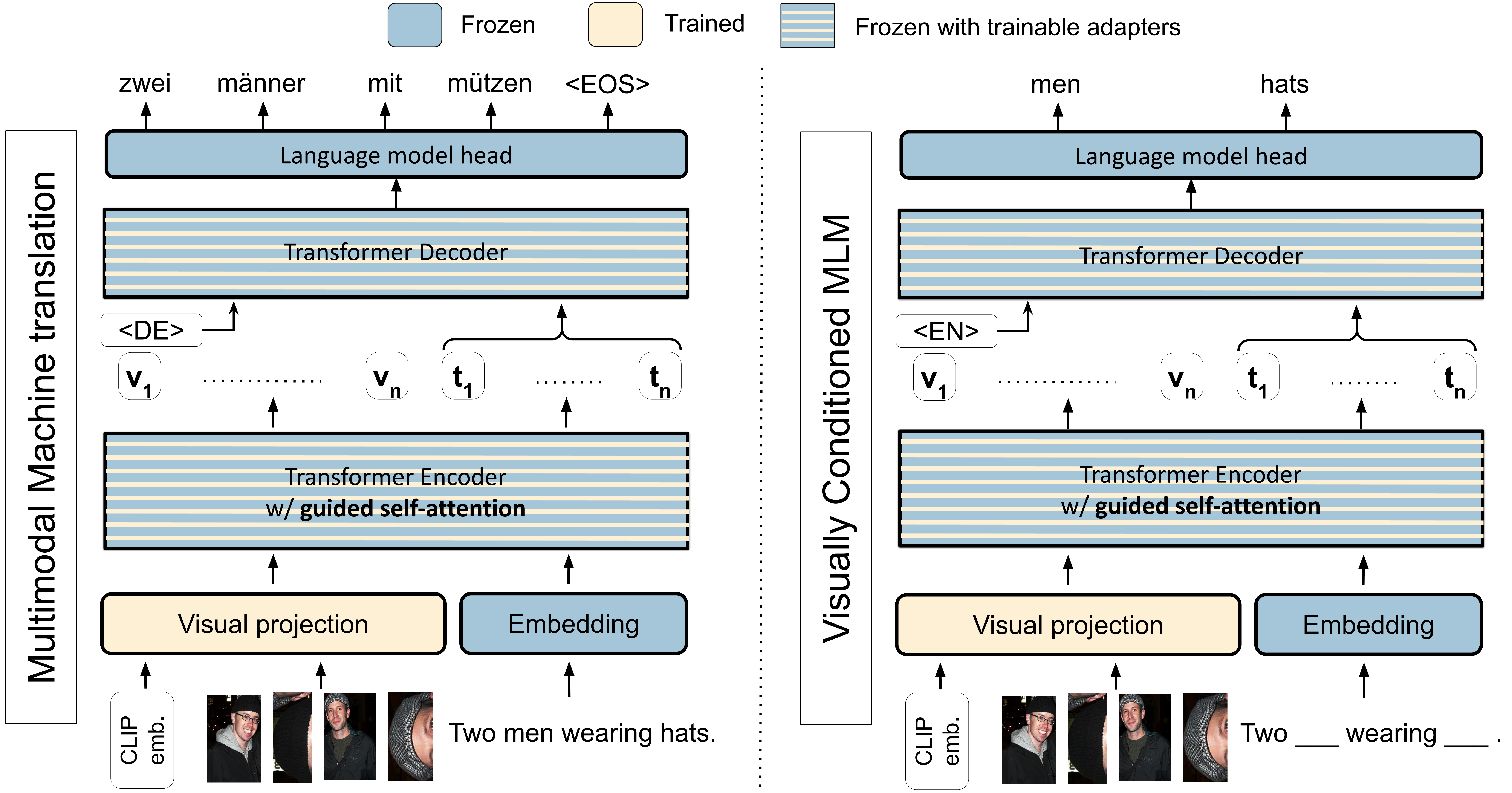}
    \caption{Overview of our approach, multimodal MT (MMT) (left) and visually-conditioned masked language modeling (VMLM) (right) objectives. We train VGAMT on both objectives jointly.}
    \label{fig:main_model}
\end{figure*}

\paragraph{Adapting pretrained LMs to multimodal inputs.}
A lot of progress has been made through the use of pretrained LMs \citep{devlin-etal-2019-bert,XLM,mbart}, often trained on raw text for text-only models or image captioning data for multimodal ones \citep{CLIP,alayrac2022flamingo,chen2022pali}.
One of the most efficient ways to learn multimodal LMs is the visually-conditioned masked language modelling (VMLM) objective \citep{chen2020uniter, lu2019vilbert, vlbert, li2020oscar, zhou2021uc2, M3P, li2019visualbert}. Inspired by the masked language modelling (MLM) objective \citep{devlin-etal-2019-bert}, it consists in randomly masking input text tokens and predicting them conditionally based on the visual features. 
A lot of interest has also been shown in lightweight modules such as adapters \citep{houlsby2019parameter} to adapt large frozen LMs to multimodal tasks \citep{eichenberg2021magma, yang2022frozenbilm, pfeiffer2021xgqa, tsimpoukelli2021multimodal,Sung2022VLADAPTERPT} in order to avoid catastrophic forgetting \citep{forgetting}. Based on these approaches, we propose to adapt a strong text-only MT model 
with lightweight modules in order to exploit the large amount of data it previously learned. 

\paragraph{Which type of visual features in MMT systems?} 
In terms of how images are represented in multimodal models, different strategies exist. Many works first proposed to incorporate global visual features from object recognition models pretrained on ImageNet \citep{image-net}, such as ResNet50 \citep{resnet}, either in the form of a single vector or a set of features \citep{doubly-calixto-2017, desmond-2017-imagination, calixto-liu-2017-incorporating, yao-wan-2020-multimodal, cuni}. More recent global features extractor such as CLIP \citep{CLIP} exist, but to our knowledge have not been used in MMT models. Extending this idea, other works focused on entities in the image and extracted bounding boxes using a pretrained Faster R-CNN \citep{faster-rcnn} in order to introduce more semantic visual information into MT \citep{gronroos-etal-2018-memad, ive-etal-2019-distilling, VTLM}. 
Recent efforts have been made to only select parts of the image that are relevant to the translation of the sentence. Some proposed to use a more selective attention mechanism between modalities \citep{Liu2021GumbelAttentionFM,ye-etal-2022-noise}, while others suggested extracting other types of visual features \citep{EMMT, Fang2022NeuralMT}. Based on this, \citet{GRAPH-MMT} decided to exploit local image-text correspondences in their model Graph-MMT. Similar to their approach, we use a simpler method to extract relevant visual features, using the output queries from a state-of-the-art free-form text object detector MDETR \citep{MDETR} as our local visual features (in addition to global features from CLIP). 

\section{Our approach: VGAMT}

The two main aims of our approach are to (i)~exploit a maximum available data (not just multimodal parallel text data) and to (ii)~provide an effective way to combine image and text modalities.
Our approach, shown in Figure~\ref{fig:main_model}, consists in taking a strong text-only MT model\footnote{In practice, our starting point is mBART, which we fine-tune on a large parallel corpus (see Section~\ref{sec:experimental} for more details).} and adapting it to multimodal MT.
To adapt this strong text-only model to multimodal inputs, we add several lightweight modules---bottleneck adapters \citep{houlsby2019parameter} and linear visual projection layers---to the otherwise frozen initial model. The bottleneck adapters are lightweight linear layers introduced after each attention block and each feed-forward layer to project embeddings down before projecting them up. 

\begin{figure}[!t]
    \centering
    \includegraphics[width=.94\linewidth]{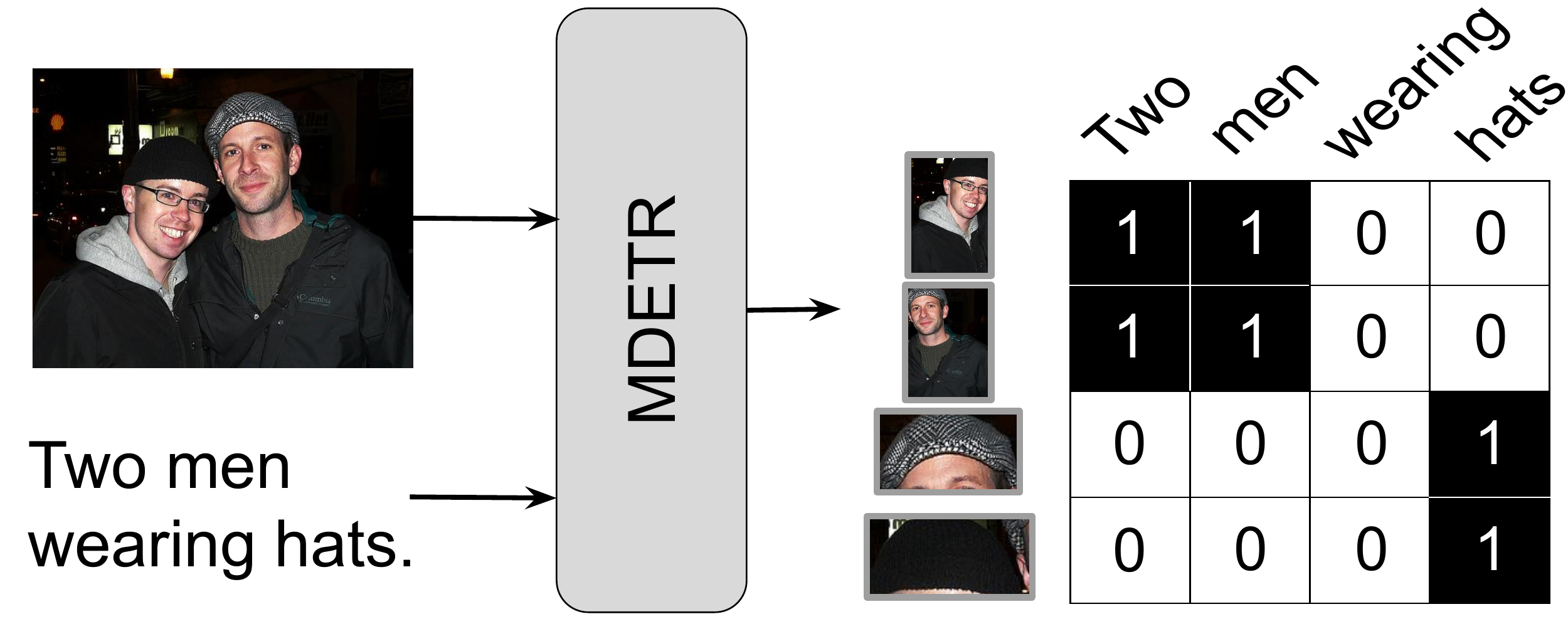}
    \caption{Example of an MDETR alignment matrix.}
    \label{fig:mdetr_example}
\end{figure}

In terms of representing visual information, we choose to use two types of representation. We concatenate local (MDETR) features and global (CLIP) features to the text inputs. We choose to use global features too, since the source sentence can describe more general aspects of the image than mere objects (such as scenes). We jointly train the non-frozen parts of our model on two distinct objectives: multimodal MT (MMT) and visually-conditioned masked language modelling (VMLM), as described in Section~\ref{sec:vmlm}.
We also introduce a guided self-attention to exploit image information in a straightforward manner (see Section~\ref{sec:guided-self-attention}) in the encoder (while the decoder uses regular self- and cross-attentions and can only attend to embeddings related to text positions). We call our approach Visually Guided and Adapted Machine Translation (VGAMT). 

\subsection{Combining training objectives} \label{sec:vmlm}

As shown in Figure~\ref{fig:main_model}, we jointly train VGAMT on two objectives: visual masked language modelling (VMLM) and multimodal MT (MMT). VMLM (resp. MMT) consists in predicting masked tokens (resp. translating the sentence) conditioned on the image.\footnote{During training, we randomly draw batches from a parallel multimodal dataset (for MMT) and a monolingual multimodal one (for VMLM) with equal probability.} 
The use of the VMLM objective in addition to MMT ensures that the model does not learn to ignore the visual inputs when translating (since Multi30k is mainly composed of very standard and unambiguous parallel sentences). 
We make sure to mask a high percentage (25\%) of the text inputs so that the model is forced to attend to the image when producing translations.

\subsection{Guided self-attention}\label{sec:guided-self-attention}

The backbone of VGAMT is an encoder-decoder MT model, in which image features are concatenated to textual input embeddings and shared self-attention is used over the two input modalities (see Figure~\ref{fig:main_model}). Instead of using full self-attention \citep{VTLM} (connections between all image parts and all text tokens), we introduce guided self-attention.
Guided self-attention consists in masking irrelevant connections between text and image representations; each text (resp. image) embedding can attend to itself and all other text (resp. image) positions, but can only attend to image (resp. text) positions conditioned on pre-extracted text-image alignments. We obtain these alignments (in the form of a cross-modal correspondence matrix) using MDETR \citep{MDETR}, which detects image regions and corresponding text spans based on a free-form text (see Figure~\ref{fig:mdetr_example} and Appendix~\ref{app:visual-features} for more details).

Concretely, let Q, K and V denote the learnable query, key and value parameters of a standard self-attention mechanism. Attention can be defined as $\text{Attention}(Q, K, V) = A \cdot V$, where the attention matrix $A=(a_{ij})$ is defined as
$A = \text{softmax}\left(QK^T/\sqrt{d_k}\right)$,
where $d_k$ is the dimension of the key vector, i.e.:
\begin{align}
a_{ij} = \frac{ e^{Q_i K_j^T/\sqrt{d_k}} }{  \sum_{l} e^{Q_i K_l^T/\sqrt{d_k}} }
    \label{eq_standard_sa}
\end{align}

The idea behind our guided self-attention mechanism is that we want to allow subwords to attend to all subwords, all bounding boxes to attend to all bounding boxes, but to only allow cross-modal attention between a subword and bounding boxes that are linked by MDETR (see Figure~\ref{fig:mdetr_example}). We therefore define a binary masking matrix $C = (c_{ij})$ where (i)~$c_{ij} = 1$ if indices $i$ and $j$ correspond to embeddings coming from the same modality, and (ii)~$c_{ij}$ is provided by the MDETR matrix otherwise: it is $1$ if MDETR has created a link between subword (resp.~bounding box) $i$ and bounding box (resp.~subword) $j$. Once this {\em guiding} matrix $C$ is defined, we can replace the standard attention (\ref{eq_standard_sa})
with our guided attention:
\begin{align}
    a_{ij} = \frac{ c_{ij}  e^{Q_i K_j^T/\sqrt{d_k}} }{  \sum_{l} c_{il} e^{Q_i K_l^T/\sqrt{d_k}} }.
    \label{eq_guided_sa}
\end{align}

The main advantage of guided self-attention over full self-attention is that the model does not have to learn to ignore irrelevant text-image correspondences since alignments are introduced as a prior.

\section{Contrastive Multilingual Multimodal Translation Evaluation (CoMMuTE)}

\begin{figure}[!ht]
    \centering
    \includegraphics[width=0.85\linewidth]{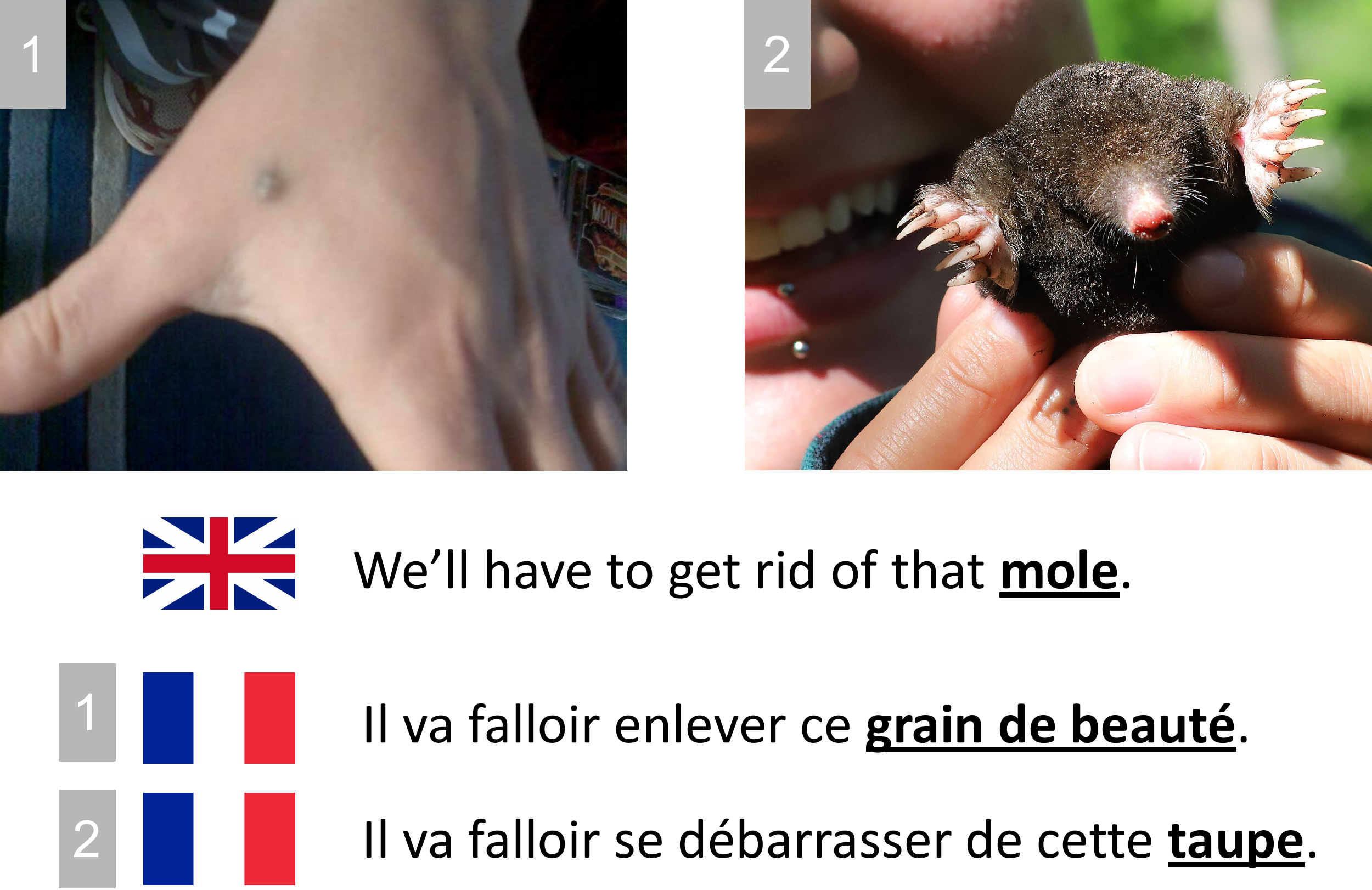}
    \caption{Example from CoMMuTE. 
    The English word `mole' can refer either to `a small dark, raised lump on a the skin' (1) or `a small burrowing mammal' (2).}
    \label{fig:commute_example}
\end{figure}

To overcome the flaws of existing benchmarks (see Section~\ref{data}), we introduce CoMMuTE, a Contrastive Multilingual Multimodal Translation Evaluation dataset\footnote{CoMMuTE is distributed under Creative Commons Attribution Share Alike 4.0 International license.}. It is composed of 155 lexically ambiguous sentences in English, each associated with two translations corresponding to two of the possible meanings of each sentence and two images that determine which of the translations is correct. It covers English$\rightarrow$French, English$\rightarrow$German and English$\rightarrow$Czech. 
An example is given in Figure~\ref{fig:commute_example}. 

\paragraph{Data collection.} 
The test set contains 155 ambiguous sentences constructed around 155 lexically ambiguous words: 29 of the examples are from \citet{bawden-etal-2018-evaluating}, and we created the  remaining ones.\footnote{We could not take the entirety of the examples in \citep{bawden-etal-2018-evaluating} as some examples were not adapted to disambiguation using visual (as opposed to linguistic) context.} We collected two images for each sentence under Creative Commons license (either Google Images or our own photos), so that the image illustrates without ambiguity one of the two meanings of the sentence. We do not restrict the image-text relation to be strictly descriptive (as for image captions) in order to have a more general evaluation dataset. Each sentence was translated into two possible translations (each corresponding to one of the images) by a native speaker of the target language. Appendix~\ref{sec:commute_stats} provides some basic statistics.

The idea of CoMMuTE is to use MMT models to rank each of the two translations based on image information. The perplexity of a sentence for a given model is defined as:
$PPL_{q}(y) = \prod_{i=1}^N q(y_i)^{-\frac{1}{N}}$,
where $q$ is the probability distribution output by the model, N is the sequence length and $y_1, \dots, y_N$ is the sequence of tokens. Now, let $y_1, \dots, y_{N_1}$ be the sequence of tokens of the correct translation and $y^\prime_1, \dots, y^\prime_{N_2}$ the sequence of tokens of the incorrect translation, a model makes a correct prediction if 
    $PPL_{q}(y) \leq PPL_{q}(y^\prime)$.
i.e. the model considers the correct translation more likely than the incorrect one. For each example, we rank each of the translations based on each of the images (2 comparisons per example), and report the accuracy over all the examples. As CoMMuTE is perfectly balanced, 
a text-only model will get exactly 50\% accuracy on this task{}. 

\begin{table}[ht]
\resizebox{\linewidth}{!}{
    \centering\small
    \begin{tabular}{lrrrrrr} \toprule
    & \multicolumn{2}{c}{En$\rightarrow$Fr} & \multicolumn{2}{c}{En$\rightarrow$De} & \multicolumn{2}{c}{En$\rightarrow$Cs} \\ \cmidrule(lr){2-3} \cmidrule(lr){4-5} \cmidrule(lr){6-7}
    & size & \#sents.  & size & \#sents.  & size & \#sents. \\ \midrule

    OpenSubtitles & 2.2GB & 24.2M & 1.2GB & 13.1M & 2.2GB & 24.7M \\
    Ted Talks & 108MB & 535K  & 83MB & 414K & 30MB & 158K \\
    Books & 29MB & 119K & 12MB & 47K &  \multicolumn{2}{c}{-} \\
    Wikipedia & 187MB & 769K & 493MB & 2.2M & 3.2MB & 19K \\ \midrule
    Total & 2.5GB & 25.6M & 1.8GB & 15.8M & 2.2GB & 24.9M \\ \bottomrule
    
    \end{tabular}
    }
    \caption{Parallel corpus sizes.}
    \label{tab:opus_data}
\end{table} \vspace{-5mm}


    

\section{Experiments} \label{sec:experimental}

\subsection{Text-only data} \label{sec:text-only-data}

All our experiments are based on the strong MT model mBART\footnote{mBART is pretrained on CC25 \citep{CCnet}.} \citep{mbart}, which we fine-tune on parallel text (see Table~\ref{tab:opus_data}). 
%
%
We use OpenSubtitles2018\footnote{\url{http://www.opensubtitles.org}} \citep{lison-etal-2018-opensubtitles2018}, Wikipedia \citep{wikipedia}, Ted Talks \citep{reimers-2020-multilingual-sentence-bert} and the Books datasets \citep{tiedemann-2012-parallel}.
We preprocess the data using Moses scripts \citep{koehn-etal-2007-moses}.\footnote{\texttt{remove-non-printing-char.pl}, \texttt{normalization-\\punctuation.pl}, and \texttt{clean-corpus-n.pl} (4-100 tokens).}

\subsection{Multimodal data} \label{data}

\begin{table}[h]
\resizebox{\linewidth}{!}{
    \centering
    \begin{tabular}{lccc} \toprule
         & Test2016 & Test2017 &  MSCOCO \\
         \midrule
       Ambiguous (\%)  & 21 (2.1\%) & 20 (2\%) & 6 (1.3\%) \\
       \bottomrule
    \end{tabular}}
    \caption{Number (and percentage) of ambiguous examples in the En$\rightarrow$Fr test sets.}
    \label{tab:ambiguous_numbers}
\end{table} \vspace{-2mm}
\paragraph{Multi30k.} We train our frozen MT model on the Multi30k dataset \citep{specia-2016-shared, elliott-2016-multi30k} composed of English sentences, each accompanied by an image and French, German and Czech translations. It contains 29k train, 1014 dev and 1000 test examples (Test2016). \citet{elliott-EtAl:2017:WMT} and \citet{barrault2018findings} released two additional related test sets (Test2017 and Ambiguous Coco). However, on analysis of these sets and as shown in Table~\ref{tab:ambiguous_numbers}, we found that very few examples are image-dependent (i.e.~the source sentence is ambiguous and the image is required to solve the ambiguity in the target language),\footnote{
We queried WordNet \citep{wordnet} for all nouns and verbs in the English sentences. An example was considered image-dependent if, on manual assessment, there were multiple meanings, which were (i)~compatible with the text context and (ii)~could be disambiguated using the image.} 
meaning that an MMT system is unlikely to perform better than a text-only system. 
Moreover, most of these ambiguities are semantically similar and they only cover a few multi-sense words. 
Although Ambiguous Coco \citep{elliott-EtAl:2017:WMT} is designed to be an ambiguous test set as it is built around multi-sense verbs, it was automatically created from sentences from MSCOCO \citep{mscoco} for which the textual context is often sufficient for disambiguation. 
These benchmarks remain useful to make sure MMT systems do not perform worse than text-only MT models on examples where images are not necessary to translate correctly. However, we consider them insufficient to assess how well MMT systems exploit images to improve translation. 

\paragraph{Monolingual multimodal data.}
For the VMLM objective, we train our model on the Conceptual Captions (CC) dataset \citep{sharma-etal-2018-conceptual} composed of 3.3M\footnote{At the time of writing, we were able to collect $\sim$2M images and trained models on this subset.} images aligned with English text. 

\subsection{Implementation details}

For all our experiments, we use the mBART implementation from Hugging Face \citep{wolf-etal-2020-transformers}. Experiments with adapters used bottleneck adapters \citep{houlsby2019parameter} with a reduction factor of 8 and ReLU activation \citep{relu}. We use the implementation provided by adapter-transformers \citep{pfeiffer2020AdapterHub}. We use a batch size of 512, the Adam optimiser \citep{kingma2014adam} with $\beta_1=0.9$, $\beta_2=0.99$ and a learning rate of $10^{-4}$ for En$\rightarrow$Fr and $10^{-5}$ for En$\rightarrow$\{De,Cs\}. We also applied 0.1 label smoothing \citep{label-smoothing} during training. We selected our final model according to the best BLEU score \citep{papineni-etal-2002-bleu} on the Multi30k dev set after at least one full pass over the Multi30k and Conceptual Captions training sets. We ran each experiment 3 times with different seeds and report the average BLEU\footnote{computed with the default parameters of sacreBLEU v2.0.0 from \href{https://github.com/mjpost/sacrebleu}{https://github.com/mjpost/sacrebleu}.} \citep{papineni-etal-2002-bleu} and COMET \citep{COMET} scores\footnote{Using the \texttt{wmt20-comet-da} model.} and the standard errors. We also report METEOR scores \citep{meteor-2005} in Appendix~\ref{app:meteor-scores}. All experiments were carried out on 8 NVIDIA V100 GPUs for $\sim$15h. 

\subsection{Baselines} We consider several text-only and multimodal baselines. All baselines except the MT models fine-tuned from mBART were trained from scratch with the original codebases and features released by the papers' authors. Models trained on the (multimodal) MT objective only where trained on Multi30k, while models jointly trained on the (multimodal) MT and (V)MLM objectives were trained on Multi30k and Conceptual Captions.

\paragraph{Text-only.} We trained a text-only Seq2seq Transformer \citep{TRANSFORMER} from scratch and a text-only Seq2Seq Transformer initialised from TLM weights \citep{XLM}. We refer to these models as Vanilla MT and TLM + MT respectively. We also trained several MT models initialised from pretrained mBART \citep{mbart} 
and which we fine-tuned 
on parallel data \citep{lison-etal-2018-opensubtitles2018, wikipedia}. We refer to these models as mBART + MT. `\textit{w/ adapters}' specifies that the model's weights are frozen except bottleneck adapters \citep{houlsby2019parameter}.
    
\paragraph{Multimodal.} We trained several state-of-the-art multimodal MT models: Graph-MMT \citep{GRAPH-MMT}, Gated Fusion \citep{good-for-misconceived-reasons-2021} and a {} Seq2Seq Transformer trained from VTLM weights \citep{VTLM} (hereafter {} VTLM + MMT).

\subsection{Results and Analysis}

\begin{figure}[!h]
    \centering
    \includegraphics[width=.95\linewidth]{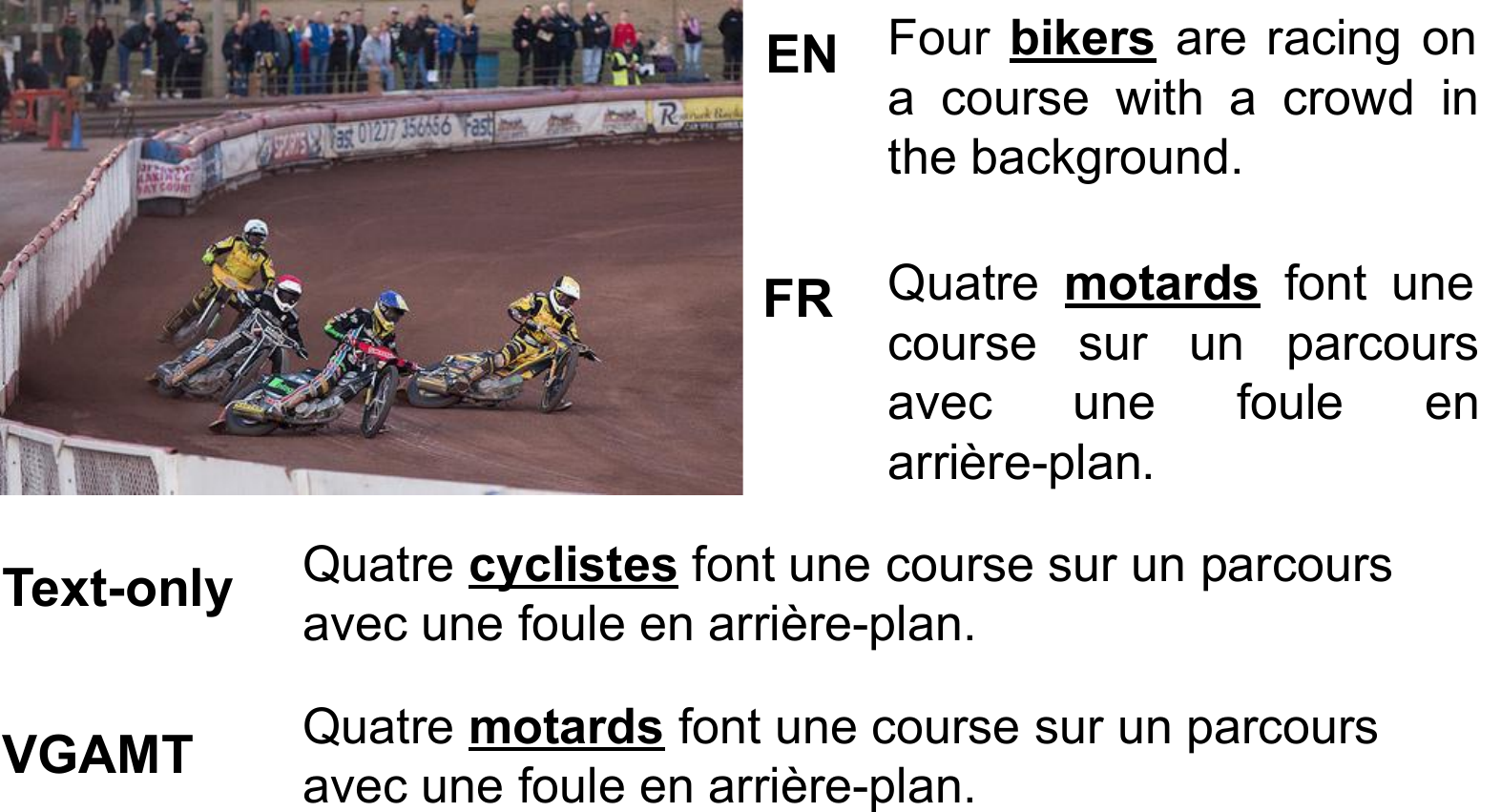}
    \caption{Example from  Test2017 with English  \textit{biker}, translated as French  `cycliste' (cyclist) or `motard' (motorcyclist). VGAMT succeeds where the baseline fails.}
    \label{fig:biker_example}
\end{figure}

\begin{figure}[!b]
    \centering
    \includegraphics[width=0.85\linewidth]{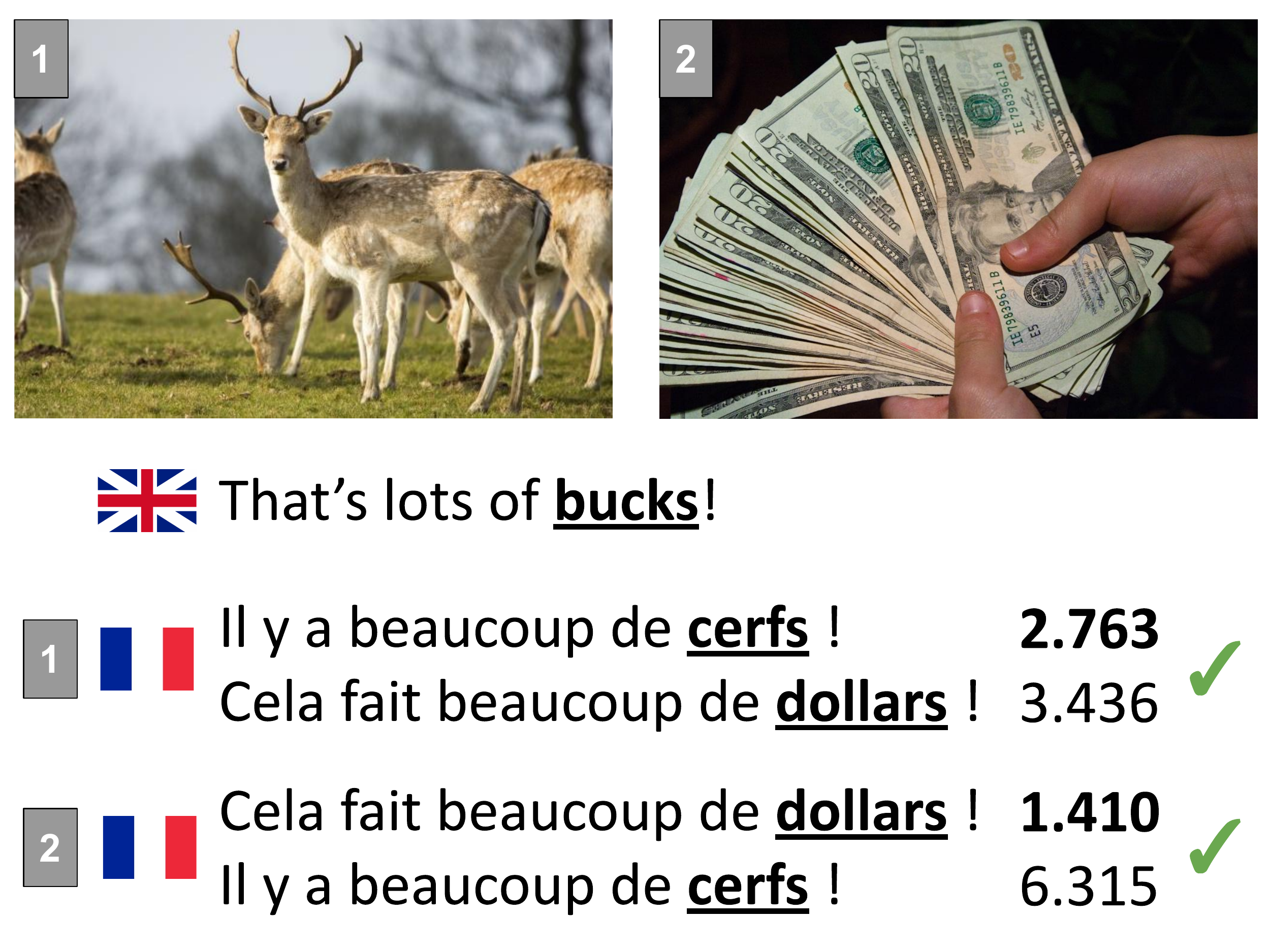}
    \caption{VGAMT Perplexity scores on a CoMMuTE example, illustrating that it is able to correctly rank each of the French translations of ambiguous English \textit{bucks} `male deer or dollars' when conditioning on the image. 
    }
    \label{fig:bucks_example}
\end{figure}

\begin{table*}[ht]
\resizebox{\linewidth}{!}{
    \centering\small
    \begin{tabular}{lccccccccc} \toprule
    \multicolumn{10}{c}{\bf En$\rightarrow$Fr} \\\midrule 
    & & \# trainable & \multicolumn{2}{c}{Test2016} & \multicolumn{2}{c}{Test2017} & \multicolumn{2}{c}{MSCOCO} & CoMMuTE \\ 
    Model & Objectives & params & BLEU & COMET & BLEU & COMET & BLEU & COMET & Accuracy \\ \midrule
    
    \multicolumn{10}{c}{\bf Text-only Machine Translation} \\ 
    \midrule
    
    Vanilla MT* 
    & NMT  & 4.0M & 59.4 {\scriptsize ±0.2} & 0.711 {\scriptsize±0.004} & 51.6 {\scriptsize±0.2} & 0.568 {\scriptsize±0.009} & 41.2 {\scriptsize±0.4} & 0.403 {\scriptsize±0.005} & \textit{50.0} \hphantom{\scriptsize±0.0} \\ 

    TLM + MT* 
    & NMT & 42M & 62.0 {\scriptsize ±0.1} & 0.795 {\scriptsize±0.002} & 54.2 {\scriptsize±0.2} & 0.681 {\scriptsize±0.002} & 43.6 {\scriptsize±0.2} & 0.542 {\scriptsize±0.009} & \textit{50.0} \hphantom{\scriptsize±0.0} \\

    mBART + MT* 
    & - & - & 49.0 \hphantom{\scriptsize±0.0} & 0.819 \hphantom{\scriptsize±0.000} & 48.1 \hphantom{\scriptsize±0.0} & 0.779\hphantom{\scriptsize±0.000} & 47.0 \hphantom{\scriptsize±0.0} & 0.733 \hphantom{\scriptsize±0.000} & \textit{50.0} \hphantom{\scriptsize±0.0} \\

    mBART + MT* \textit{w/ adapters} 
    & NMT + MLM & 12.6M & \textbf{67.2} {\scriptsize±0.3} & \textbf{0.971} {\scriptsize±0.005} & 61.5 {\scriptsize±0.3} & 0.918 {\scriptsize±0.004} & \textbf{51.5} {\scriptsize±0.7} & \textbf{0.832} {\scriptsize±0.006} & \textit{50.0} \hphantom{\scriptsize±0.0} \\ \midrule
    
    \multicolumn{10}{c}{\bf Multimodal Machine Translation} \\ 
    \midrule

    Graph-MMT* 
    & MMT  & 4.0M &  58.9 {\scriptsize±0.5} &  0.705 {\scriptsize±0.004} &  51.5 {\scriptsize±0.2} &  0.589 {\scriptsize±0.005} &  41.0 {\scriptsize±0.6} &  0.387 {\scriptsize±0.013} & 50.2 {\scriptsize±3.5}\\

    Gated Fusion* 
    & MMT  & 2.8M & 58.7 {\scriptsize ±0.3} & 0.707 {\scriptsize ±0.002} & 50.8 {\scriptsize±0.7} & 0.580 {\scriptsize ±0.011} & 40.4 {\scriptsize±0.4} & 0.394 {\scriptsize ±0.013} & 50.0 {\scriptsize±0.8}\\ 
    
    VTLM + MMT* 
    & MMT  & 44M & 61.4 {\scriptsize ±0.2} & 0.783 {\scriptsize±0.005} & 53.6 {\scriptsize±0.1} & 0.672 {\scriptsize±0.005} &  43.4 {\scriptsize±0.3} & 0.500 {\scriptsize±0.006} & 50.1 {\scriptsize±0.3} \\
    
    \textbf{VGAMT} (\textit{ours}) & MMT + VMLM & 13.2M & \textbf{67.2} {\scriptsize ±0.1} & 0.968 {\scriptsize ±0.002} & \textbf{61.6} {\scriptsize±0.1} & \textbf{0.921} {\scriptsize ±0.002} & 51.1 {\scriptsize±0.6} & 0.811 {\scriptsize±0.003} & \textbf{67.1} {\scriptsize±0.7} \\ \midrule

    \multicolumn{10}{c}{\bf En$\rightarrow$De} \\\midrule 
    
    \multicolumn{10}{c}{\bf Text-only Machine Translation} \\ 
    \midrule
    
    Vanilla MT* 
    & NMT  & 4.1M & 38.5 {\scriptsize ±0.3} & 0.394 {\scriptsize±0.005} & 30.3 {\scriptsize±0.5} & 0.259 {\scriptsize±0.012} & 27.8 {\scriptsize±0.4} & 0.092 {\scriptsize±0.018} & \textit{50.0} \hphantom{\scriptsize±0.0} \\ 

    TLM + MT* 
    & NMT & 42M & 40.0 {\scriptsize ±0.2} & 0.457 {\scriptsize±0.006} & 31.5 {\scriptsize±0.1} & 0.341 {\scriptsize±0.002} & 29.4 {\scriptsize±0.3} & 0.152 {\scriptsize±0.015} & \textit{50.0} \hphantom{\scriptsize±0.0} \\

    mBART + MT* 
    & - & - & 36.2 \hphantom{\scriptsize±0.0} & 0.595 \hphantom{\scriptsize±0.000} & 32.3 \hphantom{\scriptsize±0.0} & 0.506 \hphantom{\scriptsize±0.000} & 27.6 \hphantom{\scriptsize±0.0} & 0.383 \hphantom{\scriptsize±0.000} & \textit{50.0} \hphantom{\scriptsize±0.0} \\

    mBART + MT* \textit{w/ adapters} 
    & NMT + MLM & 12.6M & \textbf{43.6} {\scriptsize±0.2} & \textbf{0.697} {\scriptsize±0.003} & \textbf{38.9} {\scriptsize±0.5} & \textbf{0.664} {\scriptsize±0.002} & \textbf{36.2} {\scriptsize±0.2} & \textbf{0.574} {\scriptsize±0.004} & \textit{50.0} \hphantom{\scriptsize±0.0} \\ \midrule
    
    \multicolumn{10}{c}{\bf Multimodal Machine Translation} \\ 
    \midrule

    Graph-MMT* 
    & MMT  & 4.1M & 38.6 {\scriptsize±0.3} &  0.368 {\scriptsize±0.011} &  29.0 {\scriptsize±0.5} &  0.226 {\scriptsize±0.010} &  25.9 {\scriptsize±0.8} &  0.060 {\scriptsize±0.027} & 49.1 {\scriptsize±1.5}\\

    Gated Fusion* 
    & MMT  & 2.8M & 38.7 {\scriptsize ±0.2} & 0.378 {\scriptsize ±0.007} & 29.5 {\scriptsize±0.2} & 0.236 {\scriptsize ±0.018} & 26.6 {\scriptsize±0.3} & 0.055 {\scriptsize ±0.016} & 49.7 {\scriptsize±0.6}\\ 
    
    VTLM + MMT* 
    & MMT  & 44M & 39.4 {\scriptsize ±0.2} & 0.439 {\scriptsize±0.004} & 30.7 {\scriptsize±0.2} & 0.322 {\scriptsize±0.005} &  28.2 {\scriptsize±0.2} & 0.168 {\scriptsize±0.014} & 50.0 {\scriptsize±0.2} \\
    
    \textbf{VGAMT} (\textit{ours}) & MMT + VMLM & 13.2M & 43.3 {\scriptsize ±0.2} & 0.694 {\scriptsize±0.003} & 38.3 {\scriptsize ±0.2} & 0.653 {\scriptsize±0.005} & 35.7 {\scriptsize ±0.3}  & 0.544 {\scriptsize±0.006} & \textbf{59.0} {\scriptsize±0.5} \\ \bottomrule

    \end{tabular}
    }
    \caption{Results for En$\rightarrow$\{Fr,De\} (average of three runs). The best result is indicated in \textbf{bold}. * means the results were retrained by using the original codebase provided by the authors of the paper.}
    \label{tab:main_table_fr}
\end{table*}
\begin{table}[ht]
\resizebox{\linewidth}{!}{
    \centering\small
    \begin{tabular}{lccccc} \toprule
    \multicolumn{6}{c}{\bf En$\rightarrow$Cs} \\\midrule 
    & \multicolumn{2}{c}{Test2016} & \multicolumn{2}{c}{Test2018} & CoMMuTE \\ 
    Model & BLEU & COMET & BLEU & COMET & Accuracy \\ \midrule
    
    \multicolumn{6}{c}{\bf Text-only Machine Translation} \\ 
    \midrule
    
    Vanilla MT* 
    & 31.3 {\scriptsize ±0.0} & 0.593 {\scriptsize±0.008} &  
26.0 {\scriptsize±0.2} & 0.379 {\scriptsize±0.008} & \textit{50.0} \hphantom{\scriptsize ±0.0} \\ 

    TLM + MT* 
    & 32.6 {\scriptsize ±0.1} & 0.642 {\scriptsize±0.002} & 26.8 {\scriptsize±0.2} & 0.432 {\scriptsize±0.006} & \textit{50.0} \hphantom{\scriptsize ±0.0} \\

    mBART + MT* 
     & 32.1 \hphantom{\scriptsize ±0.0} & 0.865 \hphantom{\scriptsize ±0.000} & 29.6 \hphantom{\scriptsize ±0.0} & 0.747 \hphantom{\scriptsize ±0.000} & \textit{50.0} \hphantom{\scriptsize ±0.0} \\

    \tab \textit{w/ adapters}
    & 37.3 {\scriptsize±0.1} & \textbf{0.940} {\scriptsize±0.005} & \textbf{35.2} {\scriptsize±0.4} & \textbf{0.876} {\scriptsize±0.002} & \textit{50.0} \hphantom{\scriptsize ±0.0} \\ \midrule
    
    \multicolumn{6}{c}{\bf Multimodal Machine Translation} \\ 
    \midrule

    Graph-MMT* 
    & 30.8 {\scriptsize±0.4} &  0.562 {\scriptsize±0.011} &  24.9 {\scriptsize±0.5} &  0.344 {\scriptsize±0.011}  & 49.2 {\scriptsize±1.8}\\

    Gated Fusion* 
    & 30.8 {\scriptsize±0.4} & 0.560 {\scriptsize ±0.014} & 25.8 {\scriptsize±0.1} & 0.342 {\scriptsize ±0.008}  &  51.0 {\scriptsize±1.9}\\ 
    
    VTLM + MMT* 
    & 32.0 {\scriptsize ±0.3} & 0.621 {\scriptsize±0.010} & 26.7 {\scriptsize±0.2} & 0.419 {\scriptsize±0.015} & 50.0 {\scriptsize±0.3} \\
    
    \textbf{VGAMT} (\textit{ours}) & \textbf{37.6} {\scriptsize ±0.2} & 0.934 {\scriptsize ±0.004} & 34.2 {\scriptsize±0.1} & 0.833 {\scriptsize ±0.003} & \textbf{55.6} {\scriptsize±0.8} \\ \toprule
    
    \end{tabular}
    }
    \caption{Results for En$\rightarrow$Cs (average of three runs). Same formatting as for Table~\ref{tab:main_table_fr}.}
    \label{tab:main_table_cs}
    \vspace{-3mm}
\end{table}
\begin{table*}[ht]
\resizebox{\linewidth}{!}{
    \centering\small
    \begin{tabular}{lccccccc} \toprule
     & \multicolumn{2}{c}{Test2016} & \multicolumn{2}{c}{Test2017} & \multicolumn{2}{c}{MSCOCO} & CoMMuTE \\ 
    Model & BLEU & COMET & BLEU & COMET & BLEU & COMET & Accuracy \\ \midrule

    \multicolumn{8}{c}{\bf Text-only Machine Translation} \\ 
    \midrule

    mBART + MT* \textit{w/ adapters} & 67.2 {\scriptsize±0.3} & 0.971 {\scriptsize±0.005} & 61.5 {\scriptsize±0.3} & 0.918 {\scriptsize±0.004} & \textbf{51.5} {\scriptsize±0.7} & \textbf{0.832} {\scriptsize±0.006} & \textit{50.0} \hphantom{\scriptsize±0.0} \\
    
    \tab \textit{w/o MLM objective} & \textbf{67.7} {\scriptsize±0.3} & 0.970 {\scriptsize±0.004} & 61.5 {\scriptsize±0.1} & \textbf{0.926} {\scriptsize±0.004} & 50.3 {\scriptsize±0.4} & 0.821 {\scriptsize±0.002} & \textit{50.0} \hphantom{\scriptsize±0.0} \\ \midrule

    \multicolumn{8}{c}{\bf Multimodal Machine Translation} \\ 
    \midrule

    \textbf{VGAMT} (\textit{ours}) & 67.2 {\scriptsize ±0.1} & 0.968 {\scriptsize ±0.002} & \textbf{61.6} {\scriptsize±0.1} & 0.921 {\scriptsize ±0.002} & 51.1 {\scriptsize±0.6} & 0.811 {\scriptsize±0.003} & \textbf{67.1} {\scriptsize±0.7} \\
        
    \tab \textit{unfrozen w/o adapters} & 66.9 {\scriptsize ±0.7} & 0.965 {\scriptsize ±0.003} & 61.4 {\scriptsize±0.6} & 0.912 {\scriptsize ±0.009} & 50.3 {\scriptsize±0.7} & 0.814 {\scriptsize±0.011} & 60.5 {\scriptsize±3.8} \\
    
     \tab \textit{w/o VMLM objective} & \textbf{67.7} {\scriptsize ±0.2} & \textbf{0.976} {\scriptsize ±0.001} & 61.4 {\scriptsize±0.2} & 0.920 {\scriptsize ±0.003} & 50.5 {\scriptsize±0.0} & 0.809 {\scriptsize±0.004} & 52.0 {\scriptsize±1.2} \\ 
     
     \tab \textit{w/o guided self-attention} & 67.0 {\scriptsize ±0.2} & 0.963 {\scriptsize ±0.004} & 60.8 {\scriptsize±0.3} & 0.910 {\scriptsize ±0.006} & 50.3 {\scriptsize±0.5} & 0.792 {\scriptsize±0.004} & 64.6 {\scriptsize±1.6} \\ 

     \tab \textit{w/ pretraining (w/o co-training)} & 66.2 {\scriptsize ±0.1} & 0.950 {\scriptsize ±0.001} & 59.3 {\scriptsize±0.1} & 0.875 {\scriptsize ±0.003} & 49.2 {\scriptsize±0.2} & 0.777 {\scriptsize±0.001} & 63.3 {\scriptsize±0.5} \\
     
     \tab \textit{w/o MDETR features} & 66.7 {\scriptsize ±0.5} & 0.967 {\scriptsize ±0.004} & 61.1 {\scriptsize±0.1} & 0.912 {\scriptsize ±0.002} & 51.0 {\scriptsize±0.6} & 0.810 {\scriptsize±0.003} & 63.0 {\scriptsize±1.2} \\ 
     
      \tab \textit{w/o CLIP features} & 66.4 {\scriptsize±0.8} & 0.959 {\scriptsize±0.008} & 60.4 {\scriptsize±0.7} & 0.909 {\scriptsize ±0.002} & 51.0 
 {\scriptsize±0.6} & 0.810 {\scriptsize±0.008} & 50.3 {\scriptsize±0.0} \\ \bottomrule
    
    \end{tabular}}
    \caption{Results of the ablation studies described in Section~\ref{sec:ablation-study} (En$\rightarrow$Fr test set). The best result is indicated in \textbf{bold}.}
    \label{tab:ablation-en-fr}
\end{table*}

\Cref{tab:main_table_fr,tab:main_table_cs} show BLEU, COMET and accuracy scores for all models compared on several En$\rightarrow$\{Fr,De,Cs\} test sets including CoMMuTE. An initial observation is that the text-only model is a strong baseline on the three standard benchmarks (Test2016, Test2017 and MSCOCO). As mentioned in Section~\ref{data}, most of these evaluation datasets do not need visual context to be correctly translated. Our model VGAMT is on average on par with its counterpart text-only mBART+MT \textit{w/ adapters} baseline for all Multi30k En$\rightarrow$Fr test sets, while being on average just below this baseline on En$\rightarrow$\{De,Cs\} Multi30k benchmarks. It outperforms other MMT models with a large margin due to both the effective use of textual knowledge from the frozen MT model but also guided self-attention.
Note that the scores reported for the baselines are lower than the ones reported in the original papers of the models for several reasons. First, we computed the scores on fully detokenised data to have a uniform evaluation between all models. We also report the average score from three different runs using different seeds and not the best score obtained over a single run.

More importantly, our VGAMT obtains strong improvements over both text-only baselines and state-of-the-art MMT systems on CoMMuTE; our model can use visual context to disambiguate sentences. 
This can be seen in Figure~\ref{fig:biker_example} (one of the ambiguous examples from Multi30k), where in contrast to the baseline VGAMT produces the correct translation and Figure~\ref{fig:bucks_example} (from CoMMuTE), where VGAMT correctly ranks the two translations. More examples are provided in Appendix~\ref{app:additional-examples}. We also propose to translate CoMMuTE source sentences and compare against the reference translations; the results are shown in Appendix~\ref{app:translating-commute}. 

\section{Ablation Study} \label{sec:ablation-study}

To better understand the role of VGAMT's components, we carry out several ablations for En$\rightarrow$Fr and report all results in Table~\ref{tab:ablation-en-fr}.

\paragraph{Adapters versus Fine-tuning.}
We compare the results of fine-tuning an unfrozen VGAMT model (w/o adapters) in comparison to our frozen model with adapters (VGAMT), all other things remaining equal. The unfrozen version faces a drop in scores on all test sets except Test2017. Notably, the unfrozen model's accuracy score of 60.5 on CoMMuTE is 6.6 points lower than our final VGAMT model. As well as providing a more lightweight solution that does not involve fine-tuning all parameters, using neural adapters and freezing other weights is useful in terms of performance. 

\paragraph{Impact of the VMLM objective.}
To evaluate the impact of jointly training with MMT and VMLM objectives, we train a model on the MMT without VMLM (and therefore without monolingual multimodal data). 
The MMT model trained on MMT alone obtains 52.0 on CoMMuTE, compared to 67.1 for joint training, showing that VMLM helps our model to better exploit disambiguating images.  

\paragraph{Guided self-attention.}
We study the impact of  guided self-attention between modalities by comparing against classic full self-attention. 
Guided self-attention obtains better results than full self-attention, particularly on Test2017 and MSCOCO (+0.8 BLEU, +0.015 COMET on average). It also gets better results on CoMMuTE (+2.5 points). See Appendix~\ref{app:analysis-guided-self-attention} for analysis of guided attention scores.

\paragraph{VMLM and MMT joint training.}
We compare our VMLM and MMT joint training with disjoint training where VGAMT is first pretrained on VMLM then fine-tuned on MMT instead of co-training on both VMLM and MMT. Table~\ref{tab:ablation-en-fr} shows that it results in a large drop of performance on all scores in average including 3.8 points on CoMMuTE.

\paragraph{MDETR.} 
We examine the impact of MDETR features by training a model without them.\footnote{More details are available in Appendix~\ref{app:visual-features}.\label{fn:repeat}}
The results without MDETR features are slightly lower than the full model on standard MMT benchmarks. However, the results are significantly lower on CoMMuTE (63.0±1.2 without MDETR features and 67.1±0.7 with MDETR features). This means that VGAMT benefits from MDETR features when disambiguating and translating sentences.

\paragraph{CLIP.} 
We also study the impact of CLIP features by training a model without them.\footref{fn:repeat} 
Including CLIP features gives slightly higher results on standard MMT benchmarks (+0.69 BLEU and +0.007 COMET scores on average on all benchmarks). VGAMT without CLIP features faces an extreme drop on CoMMuTE (50.3±0.00 w/o CLIP features vs.~67.1±0.7 w/ CLIP features), which shows that CLIP features are required for disambiguation.

\begin{figure}[!ht]
 \centering\small
\subcaptionbox{Variation of VMLM sampling probability with fixed 25\% masked text inputs. \label{fig:sampling-vmlm}}{\includegraphics[width=0.48\linewidth]{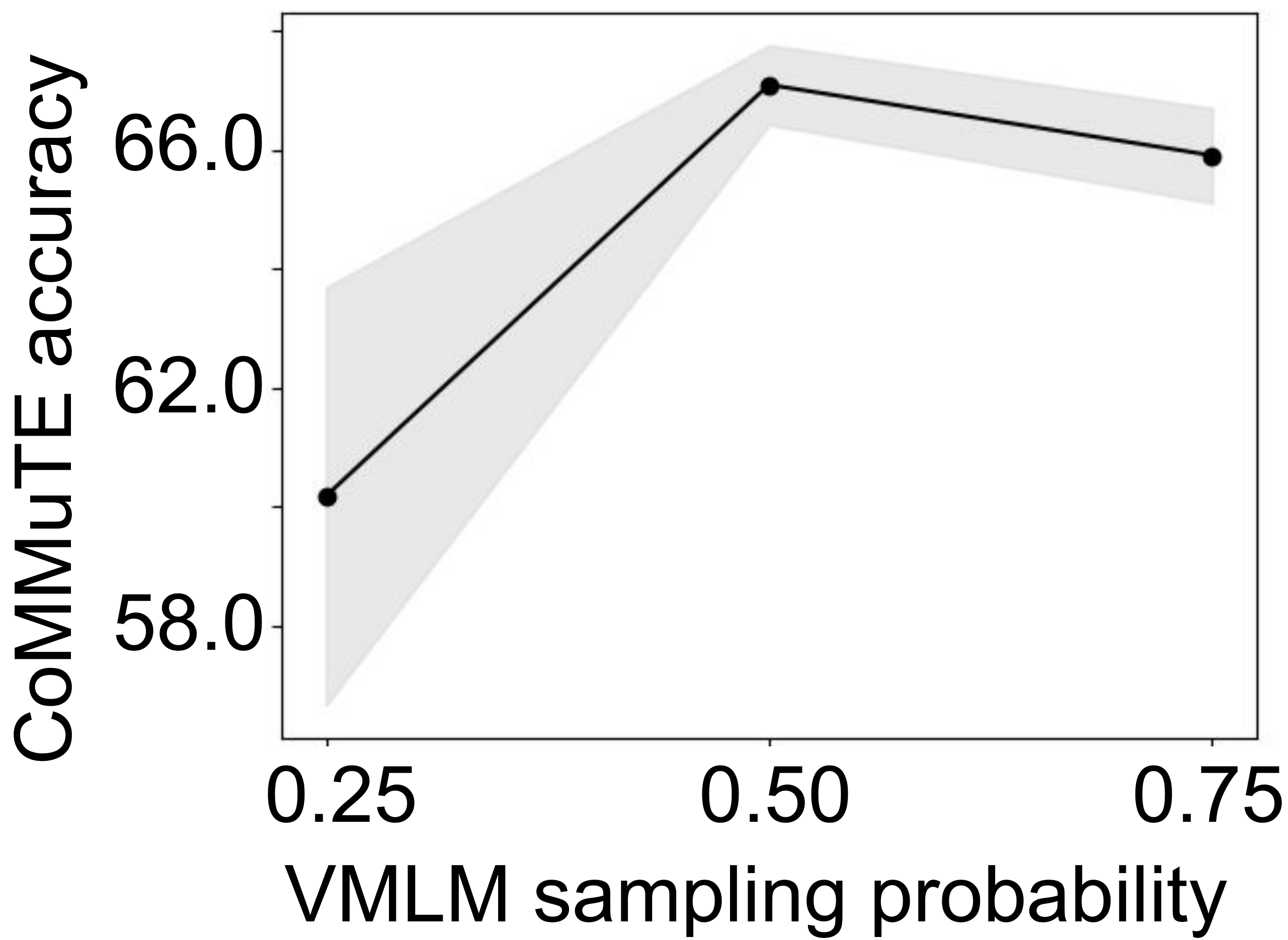}}%
\hspace{.7em}
\subcaptionbox{Variation of masked text inputs with fixed 50\% VMLM sampling probability.\label{fig:mask-text}}{\includegraphics[width=0.48\linewidth]{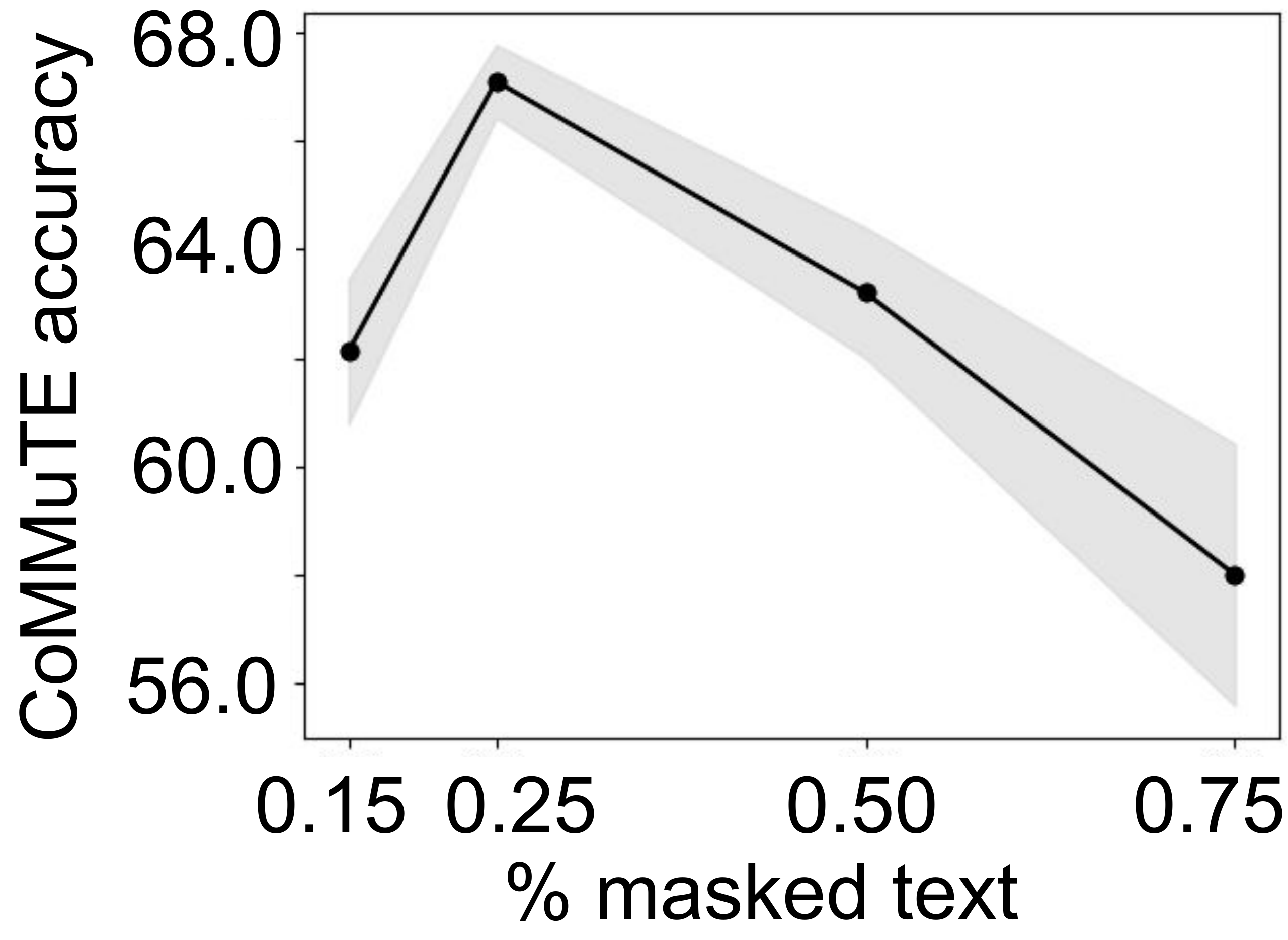}}
\caption{CoMMuTE Results comparing multiple VMLM sampling probabilities and percentage of masked text inputs. 95\% confidence interval in grey.}
\label{fig:sampling-mask}
\vspace{-5mm}
\end{figure}

\paragraph{VMLM sampling probability and degree of masking.}
We ran experiments to vary the VMLM sampling probability (see Section~\ref{sec:vmlm}) and the percentage of masked text inputs (see Figure~\ref{fig:sampling-mask} for results on CoMMuTE). For the sampling between VMLM and MMT objectives, the maximum value is reached for $p=$50\%, i.e.~equal sampling between VMLM and MMT objectives (Figure~\ref{fig:sampling-vmlm}). Similar results are obtained for $p=75\%$, i.e.~3 VMLM batches for 1 MMT batch, but the translation quality is lower. For the percentage of masking, there is a peak at 25\% masked text inputs and a constant decrease for higher values (Figure~\ref{fig:mask-text}). 

\section{Conclusion}
We propose a new MMT approach (VGAMT) based on (i)~adapting a strong text-only MT model with lightweight adapters and (ii)~introducing better use of the text and image modalities through a novel guided self-attention mechanism and joint MMT and VMLM training. 
We also introduce CoMMuTE, a contrastive test set designed to test the use of visual disambiguating context.
Results for En$\rightarrow$\{Fr,De,Cs\} show that VGAMT obtains competitive results compared with strong text-only baselines on standard benchmarks and widely outperforms these baselines and state-of-the-art MMT systems on CoMMuTE.

\section*{Limitations}
In this work, we focused on En$\rightarrow$\{Fr,De,Cs\} multimodal MT. At the time of writing, our method can only be applied for En$\rightarrow$X MMT. It is indeed necessary to have access to a modulated object detector in the source language to extract the features and the image-text relationship exploited by our model. This type of modulated object detector is only available in English for the moment. We leave the extension of our method to non-English source languages to future work. Moreover, our method requires large amount of captioning data to perform well. It is therefore computationally expensive. 

\section*{Acknowledgements}
This work was granted access to the HPC resources of IDRIS under the allocation 2022-AD011013908 and 2022-AD011012254 made by GENCI. It was also partly funded by the last four authors' chairs in the PRAIRIE institute funded by the French national agency ANR as part of the ``Investissements d'avenir'' programme under the reference ANR-19-P3IA-0001.

\bibliography{custom}
\bibliographystyle{acl_natbib}

\appendix

\section{CoMMuTE statistics}
\label{sec:commute_stats}

Some basic statistics of the CoMMuTE dataset can be found in Table~\ref{tab:commute_stats}. The source side of the dataset is always English and two translations of each of the 155 English ambiguous sentences are provided in French, German and Czech.

\begin{table}[!ht]
    \centering\small
    \begin{tabular}{lrrrr} \toprule
         & En & Fr & De & Cs \\
         \midrule
        \#unique sents. & 155 & 308 & 300 & 308 \\
        Avg. sent. length  & 6.54 & 6.90 & 6.48 & 5.07 \\
       \#unique toks & 462 & 679 & 638 & 718 \\
       \bottomrule
    \end{tabular}
    \caption{CoMMuTE statistics.}
    \label{tab:commute_stats} \vspace{-4mm}
\end{table}

\section{Visual features}\label{app:visual-features}
We use MDETR \citep{MDETR} features as our local visual features. Concretely, we extract the set of output queries features of size 64 from the MDETR decoder and introduce them as input. 

In addition, we use CLIP \citep{CLIP} features as our global visual features. More specifically, we extract the output [CLS] features of size 512 from the ViT \citep{vit} image encoder used by CLIP and introduced it as input. 

\section{Guided self-attention analysis}\label{app:analysis-guided-self-attention}

\begin{figure*}[!h]
 \centering\small
   \begin{subfigure}[b]{.25\textwidth}
 \centering
 \includegraphics[width=\linewidth]{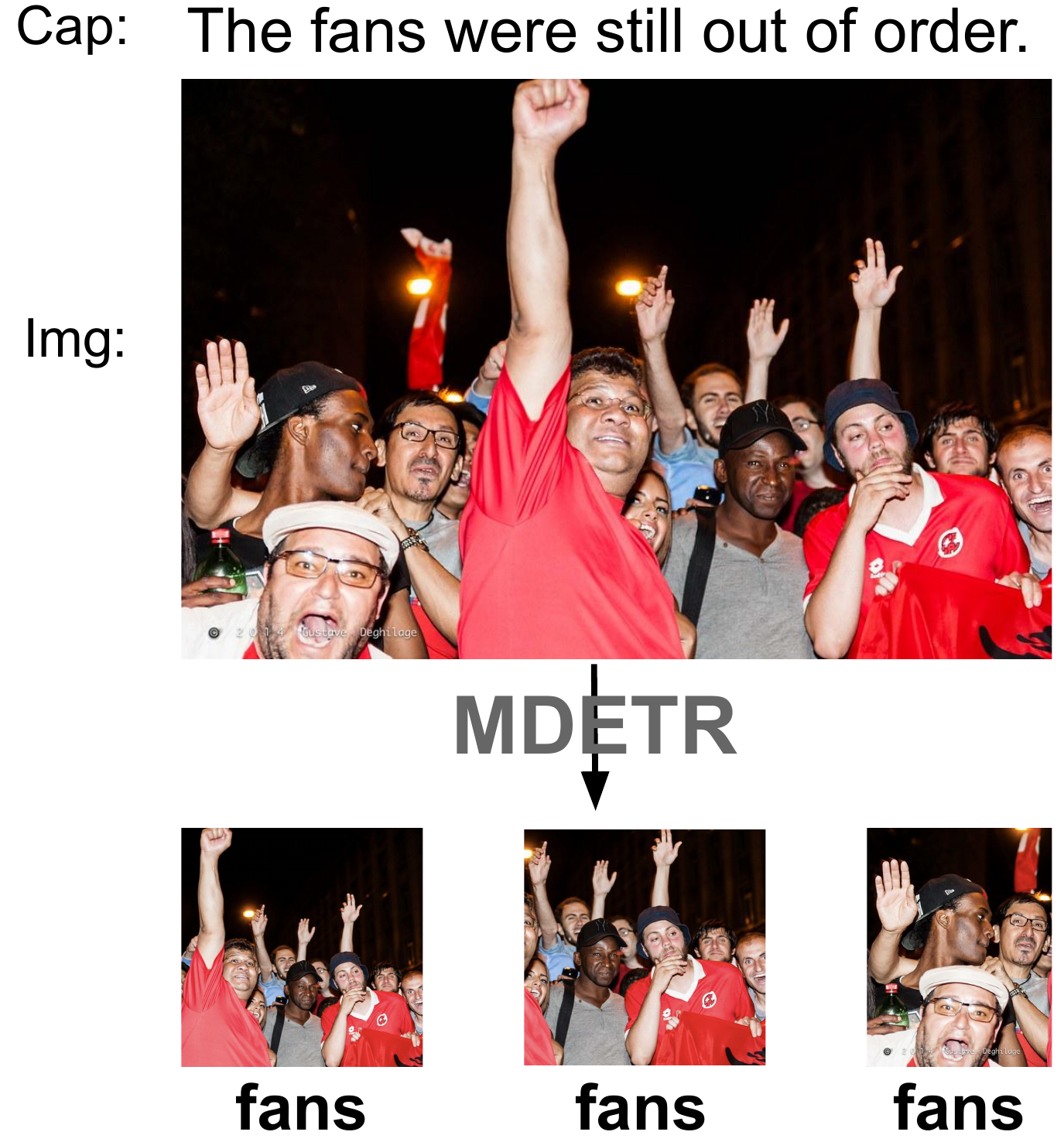}
 \caption{English sentence, its associated image and the region-text couples extracted by MDETR.}
 \label{fig:inputs_fans}
\end{subfigure}
\hspace{1em}
\begin{subfigure}[b]{.72\textwidth}
  \centering
  \includegraphics[width=\linewidth]{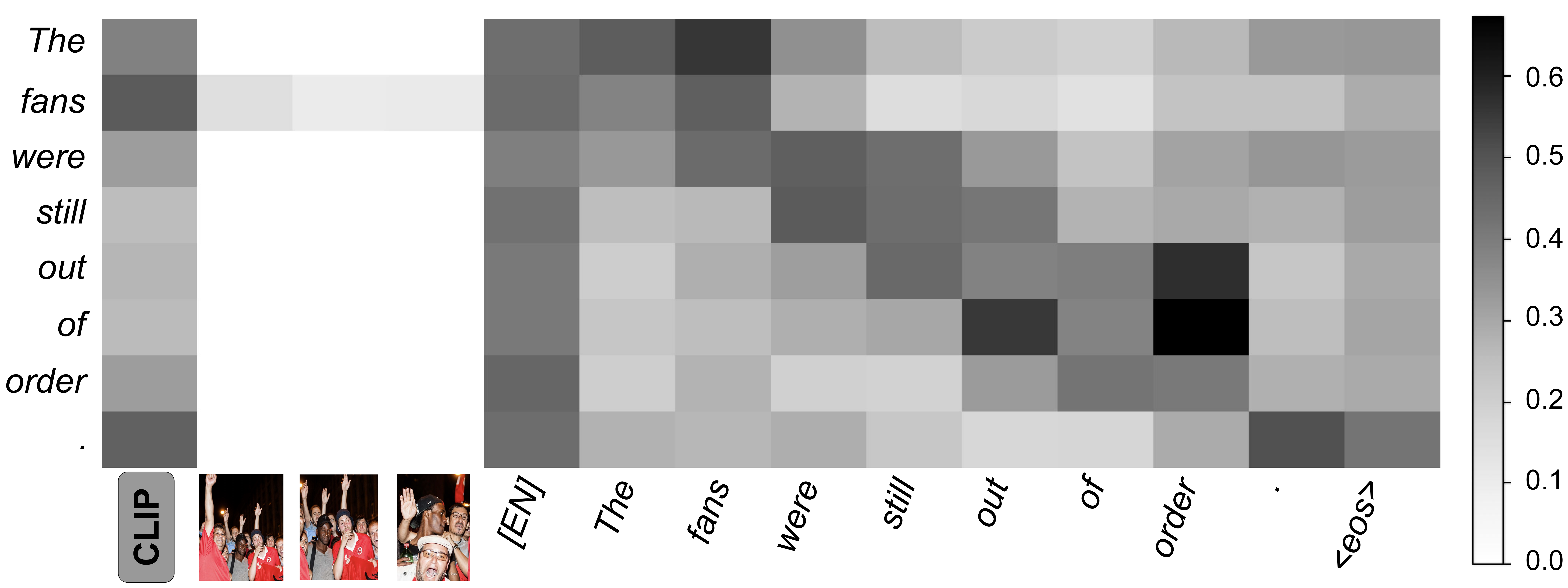}
 \caption{Normalised attention scores from the cross-modal part of our guided self-attention mechanism. Only the embeddings of the token `\textit{fans}' can attend to MDETR features. All the embeddings can attend to CLIP features.}
    \label{fig:attn_map_fans}
\end{subfigure}
\par\bigskip
\caption{Guided self-attention map for the English sentence \textit{The fans were still out of order} and its associated image. The sentence is ambiguous as English \textit{fan}  refers to `a cooling device' or `an ardent admirer'. Values are averaged over all heads and all layers. \textit{<eos>} refers to the end of sentence token.}
\label{fig:attn_fans}
\end{figure*}

We studied the values of the cross-modal part of our guided self-attention. To do so, we followed the method proposed by \citet{kobayashi-etal-2020-attention} who showed that raw attention scores $\alpha$ are meaningless and instead proposed to conduct analysis on the normalised attention scores $\lVert \alpha f \rVert$, where $\alpha$ are the raw attention scores and $f$ is the value vector in the attention mechanism. 
\begin{table*}[!ht]
\resizebox{\linewidth}{!}{
    \centering\small
    \begin{tabular}{lcccccccc} \toprule
    & \multicolumn{3}{c}{\bf En $\rightarrow$ Fr} & \multicolumn{3}{c}{\bf En $\rightarrow$ De} & \multicolumn{2}{c}{\bf En $\rightarrow$ Cs} \\ \cmidrule(lr){2-4} \cmidrule(lr){5-7} \cmidrule(lr){8-9}
    & Test2016 & Test2017  & MSCOCO & Test2016  & Test2017 & MSCOCO & Test2016 & Test2018 \\ \midrule

    \multicolumn{9}{c}{\bf Text-only Machine Translation} \\ 
    \midrule
    
    Vanilla MT* 
    & 74.4 {\scriptsize±0.1}  & 68.3 {\scriptsize±0.2} & 61.5 {\scriptsize±0.4}  & 55.0 {\scriptsize±0.2} & 46.9 {\scriptsize±0.4} & 45.3 {\scriptsize±0.3} & 30.5 {\scriptsize±0.1} & 26.5 {\scriptsize±0.1}  \\ 

    TLM + MT* 
    &  76.3 {\scriptsize±0.1} & 70.3 {\scriptsize±0.2} & 63.4 {\scriptsize±0.3} & 56.0 {\scriptsize±0.2} & 48.1 {\scriptsize±0.1} & 46.1 {\scriptsize±0.2} & 31.0 {\scriptsize±0.0} & 26.6 {\scriptsize±0.1}  \\

    mBART + MT* 
    & 68.3 \hphantom{\scriptsize±0.0} & 66.8 \hphantom{\scriptsize±0.0} & 66.4 \hphantom{\scriptsize±0.0} & 52.6 \hphantom{\scriptsize±0.0} & 48.3 \hphantom{\scriptsize±0.0} & 44.2 \hphantom{\scriptsize±0.0} & 30.7 \hphantom{\scriptsize±0.0} & 28.1 \hphantom{\scriptsize±0.0} \\

    mBART + MT* \textit{w/ adapters} 
    & \textbf{79.9} {\scriptsize±0.3} & \textbf{76.0} {\scriptsize±0.2} & \textbf{69.5} {\scriptsize±0.6} & \textbf{58.5} {\scriptsize±0.1} & \textbf{53.9} {\scriptsize±0.3} & \textbf{51.7} {\scriptsize±0.2} & \textbf{33.8} {\scriptsize±0.2} & \textbf{31.4} {\scriptsize±0.2} \\ \midrule
    
    \multicolumn{9}{c}{\bf Multimodal Machine Translation} \\ 
    \midrule

    Graph-MMT* 
     & 74.1 {\scriptsize±0.4} & 68.7 {\scriptsize±0.5} & 61.6 {\scriptsize±0.6} & 54.4 {\scriptsize±0.4} & 45.7 {\scriptsize±0.4} & 43.2 {\scriptsize±0.7} & 30.1 {\scriptsize±0.1} & 26.0 {\scriptsize±0.2} \\

    Gated Fusion* 
     & 73.1 {\scriptsize ±0.3} & 67.1 {\scriptsize ±0.5} & 60.1 {\scriptsize ±0.4} & 54.9 {\scriptsize ±0.4} & 46.2 {\scriptsize ±0.3} & 44.2 {\scriptsize ±0.4} & 28.8 {\scriptsize ±0.2} & 25.1 {\scriptsize ±0.1} \\ 
    
    VTLM + MMT* 
     & 75.9 {\scriptsize±0.1} & 69.8 {\scriptsize±0.1} & 63.3 {\scriptsize±0.2} & 55.4 {\scriptsize±0.1} & 47.7 {\scriptsize±0.1} & 45.6 {\scriptsize±0.3} & 30.6 {\scriptsize±0.1} & 26.4 {\scriptsize±0.1}\\
    
    \textbf{VGAMT} (\textit{ours})  & 79.7 {\scriptsize ±0.0} & 75.9 {\scriptsize ±0.1} & 68.9 {\scriptsize±0.4} & 58.1 {\scriptsize ±0.2}  & 53.6 {\scriptsize ±0.2} & \textbf{51.7} {\scriptsize±0.2} & 33.7 {\scriptsize ±0.1} & 30.5 {\scriptsize ±0.0}  \\ \bottomrule
    
    \end{tabular}
    }
    \caption{METEOR scores for standard En$\rightarrow$\{Fr,De,Cs\} benchmarks (average of three runs). The best result is indicated in \textbf{Bold}. * means the results were retrained by using the original codebase provided by the authors of the paper.}
    
    \label{tab:meteor-scores}
\end{table*}
Figure~\ref{fig:attn_map_fans} shows the cross-modal part of the guided self-attention map from the example displayed in Figure~\ref{fig:inputs_fans} where all the values have been averaged over all heads and all layers. In this example, the English word \textit{fans} `cooling device or ardent admirer' is ambiguous and the two meanings have different translations in French, German and Czech. Given the region-text couples extracted by MDETR (Figure~\ref{fig:inputs_fans}), only the token \textit{fans} can attend to the MDETR region features. The normalised attention scores of the embedding of the token \textit{fans} on these regions are low in comparison to the scores on the text part and on the CLIP embedding. On the contrary, all embeddings can attend to CLIP embedding and the embedding of the token \textit{fans} is the one with the highest normalised attention score with CLIP embedding.

\section{Additional examples} \label{app:additional-examples}

\begin{figure}[!h]
    \centering
    \includegraphics[width=\linewidth]{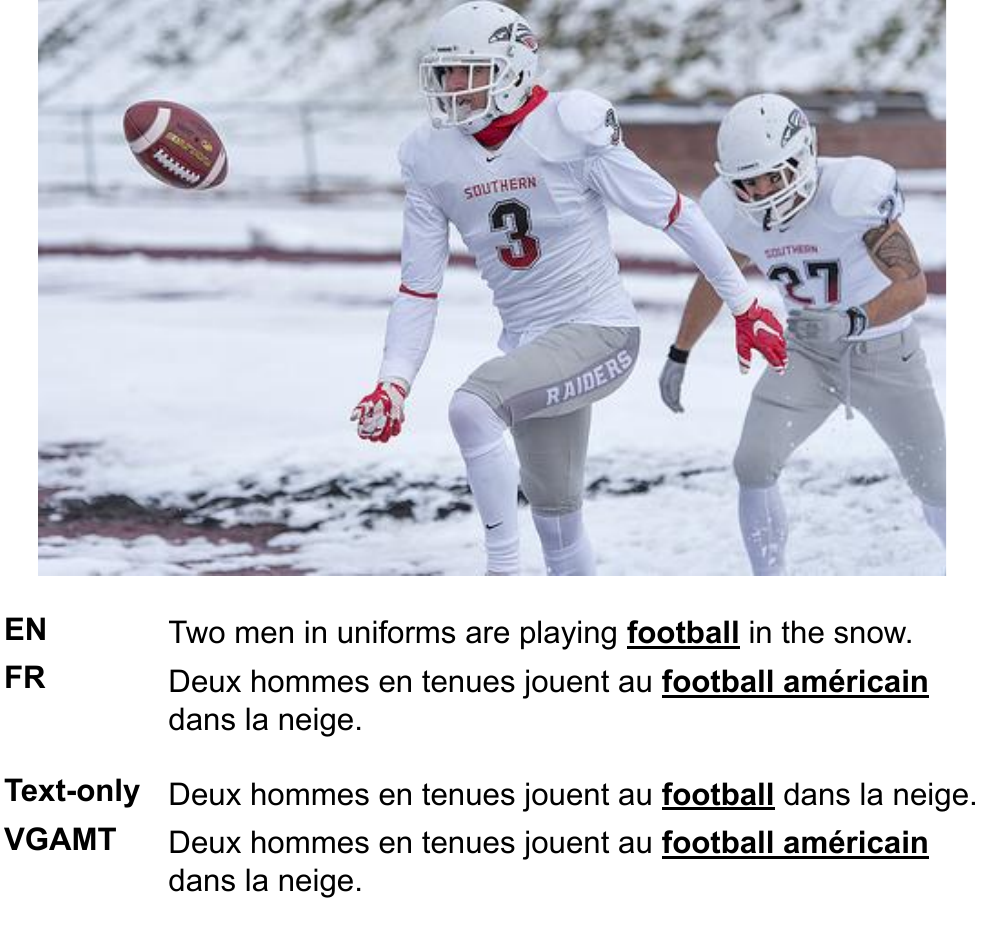}
    \caption{Example from Test2017, illustrating how VGAMT is able to exploit visual information to distinguish between the two types of football (soccer (1) and American football (2) depending on whether British or American English is used), whereas the text-only baseline produces a wrong translation.}
    \label{fig:football_example}
\end{figure}

\begin{figure*}[t]
 \centering\small
   \begin{subfigure}[b]{0.49\textwidth}
 \centering
 \includegraphics[width=.95\linewidth]{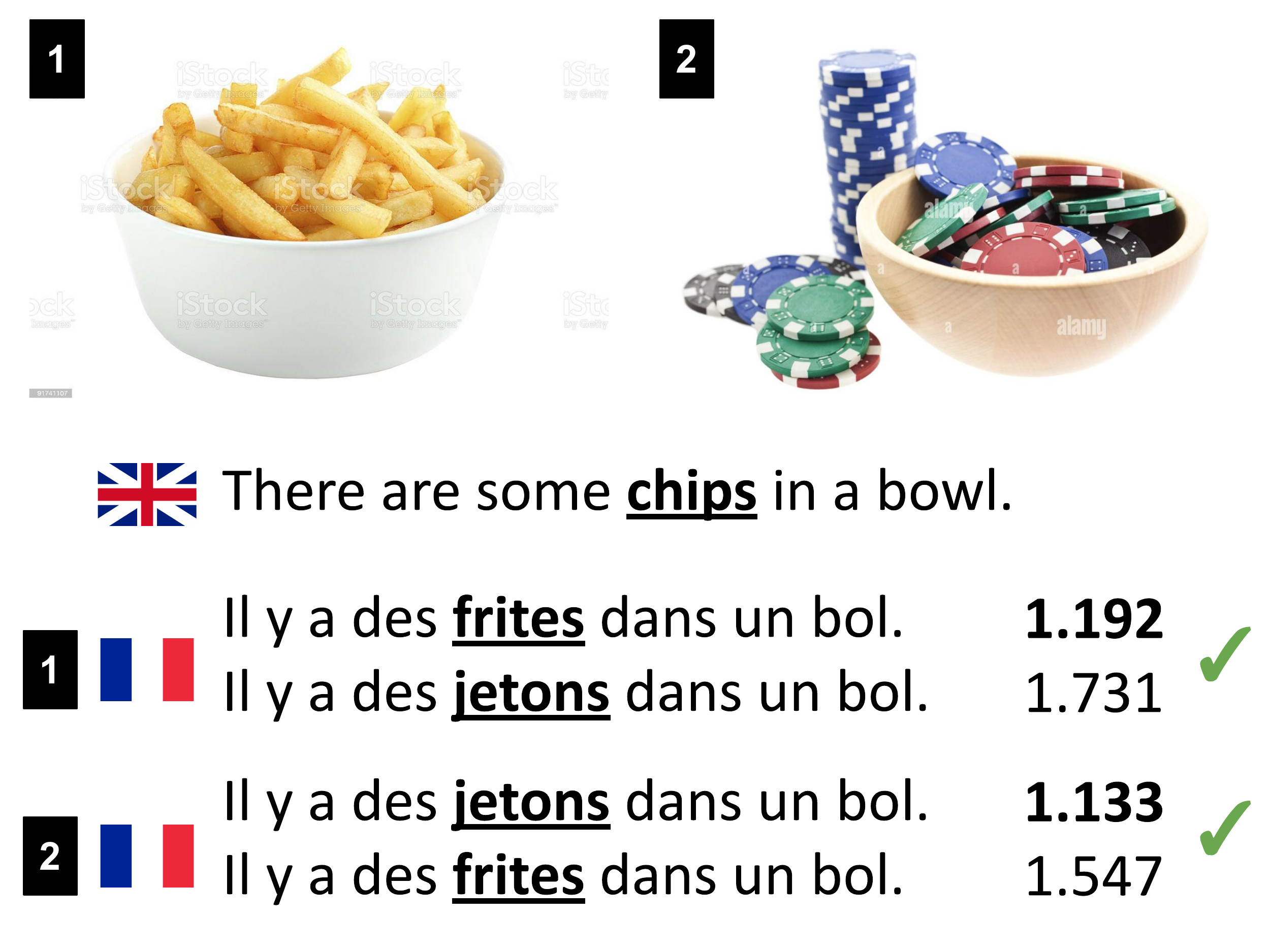}
 \caption{The English word \textit{chips} refers to `french fries' or `poker chips'.}
 \label{fig:chips_ex}
\end{subfigure}
\hfill
\begin{subfigure}[b]{0.49\textwidth}
  \centering
  \includegraphics[width=.95\linewidth]{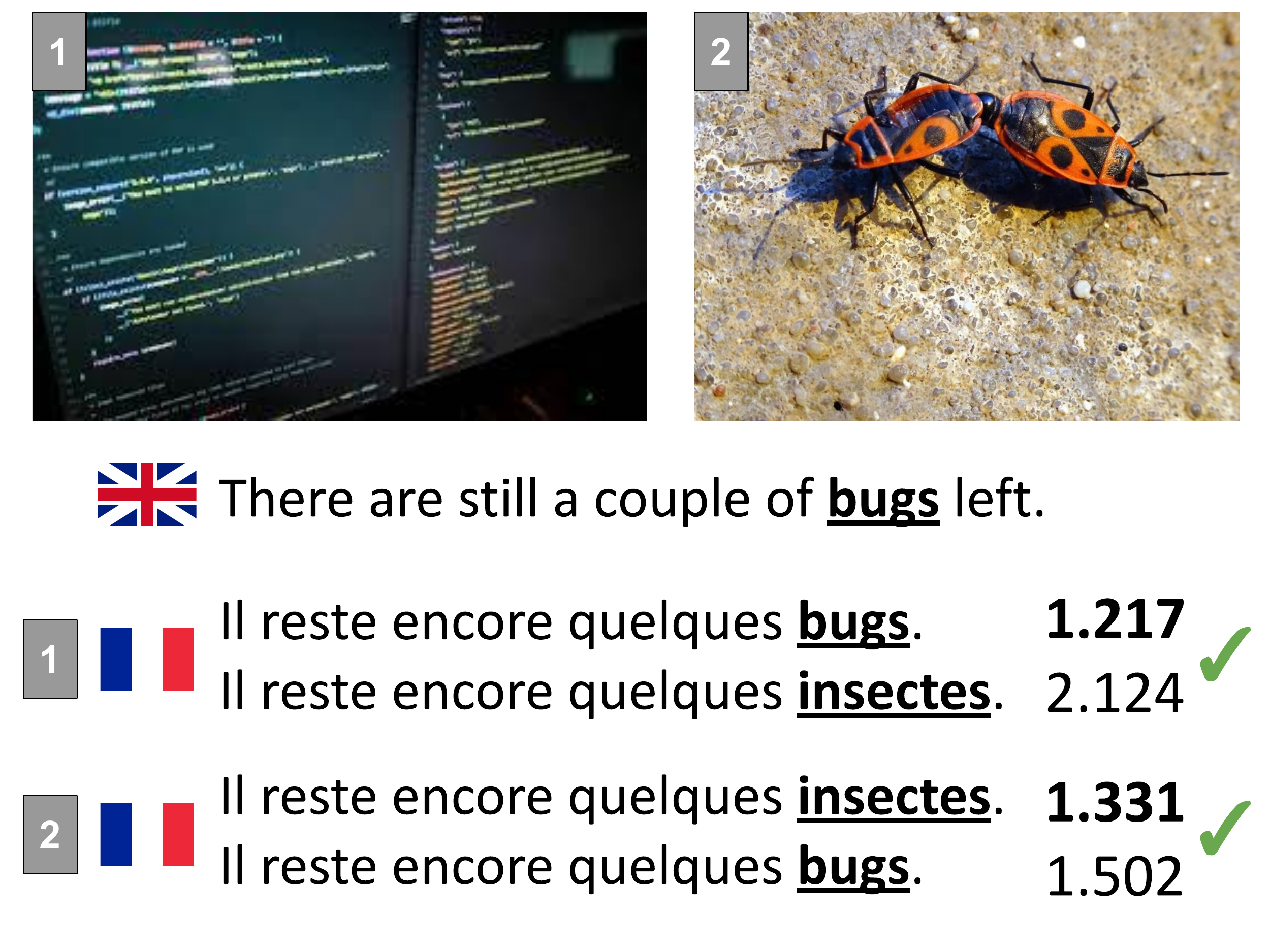}
 \caption{The English word \textit{bugs} refers to `a problem in a computer program' or `a small insect'.}
    \label{fig:bugs_ex}
\end{subfigure}
\par\bigskip 
\begin{subfigure}[b]{0.49\textwidth}
  \centering
  \includegraphics[width=.95\linewidth]{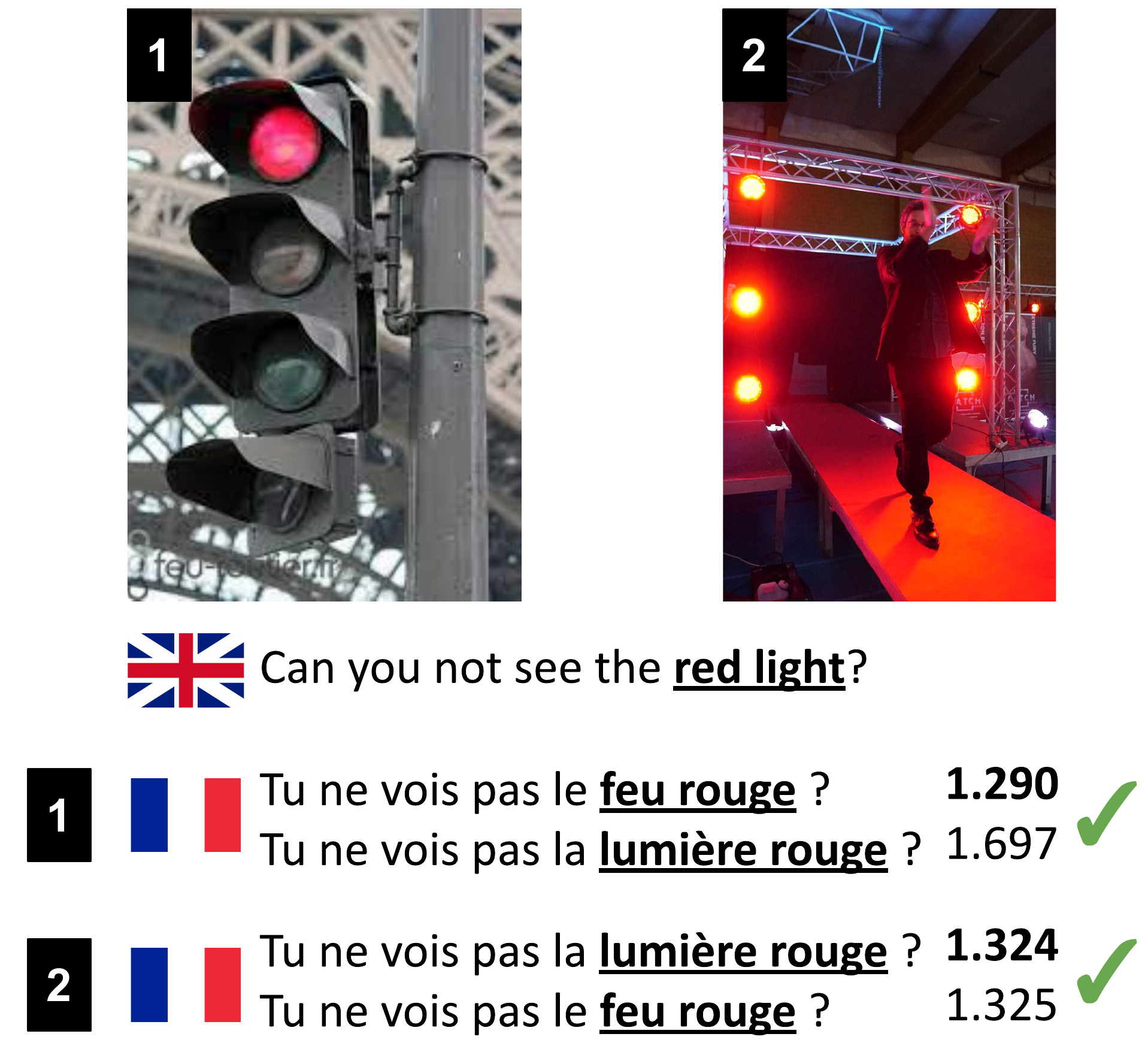}
 \caption{The English phrase \textit{red light} refers to `a traffic signal that instructs moving vehicles to stop' or `light that is red'.}
    \label{fig:lights_ex}
\end{subfigure}
\hfill
\begin{subfigure}[b]{0.49\textwidth}
  \centering
  \includegraphics[width=.95\linewidth]{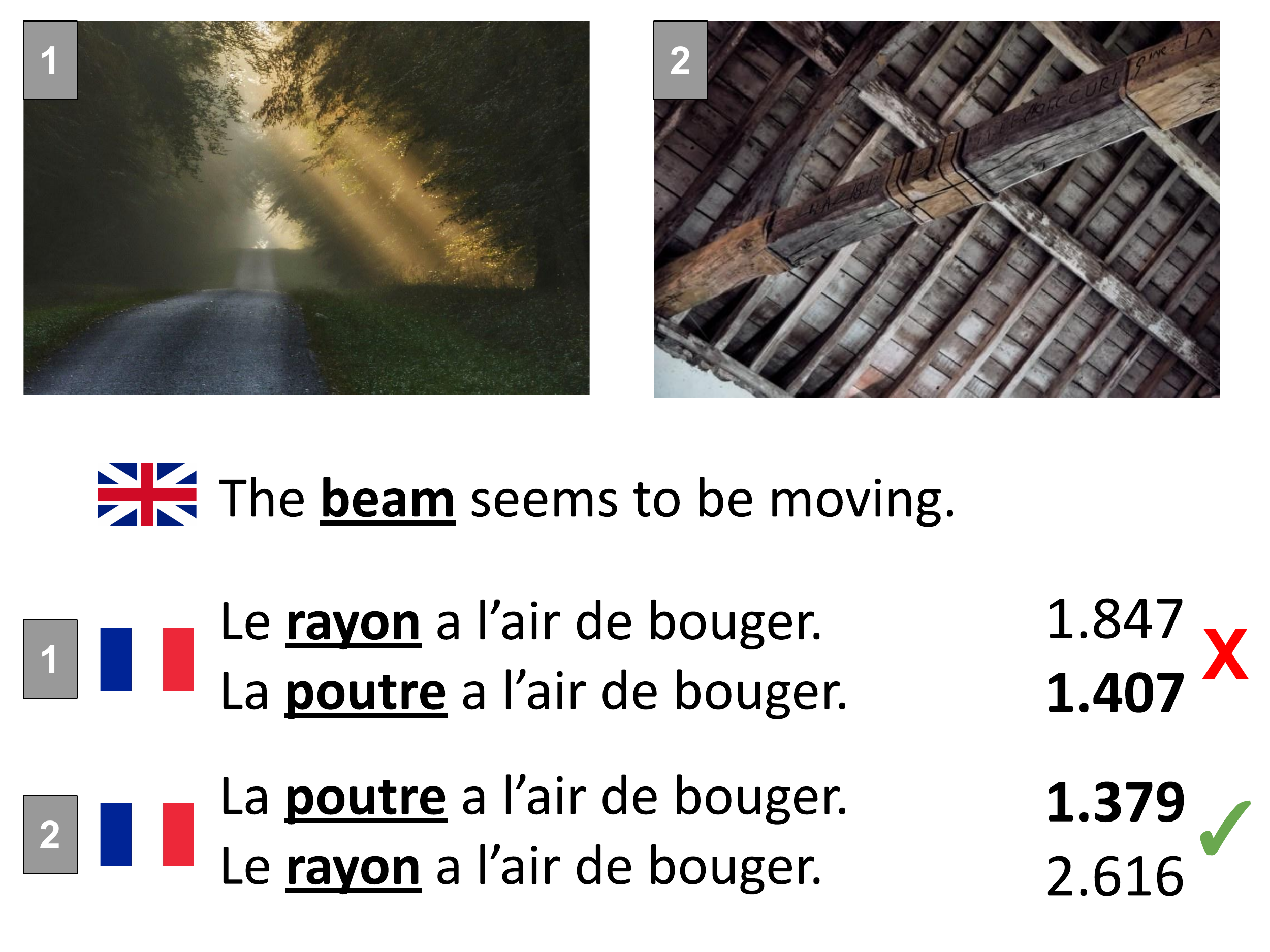}
 \caption{The English word \textit{beam} refers to `a ray of light' or `a piece of timber or metal used to support the roof'.}
    \label{fig:beams_ex}
\end{subfigure}
\par\bigskip 
\begin{subfigure}[b]{0.49\textwidth}
  \centering
  \includegraphics[width=.95\linewidth]{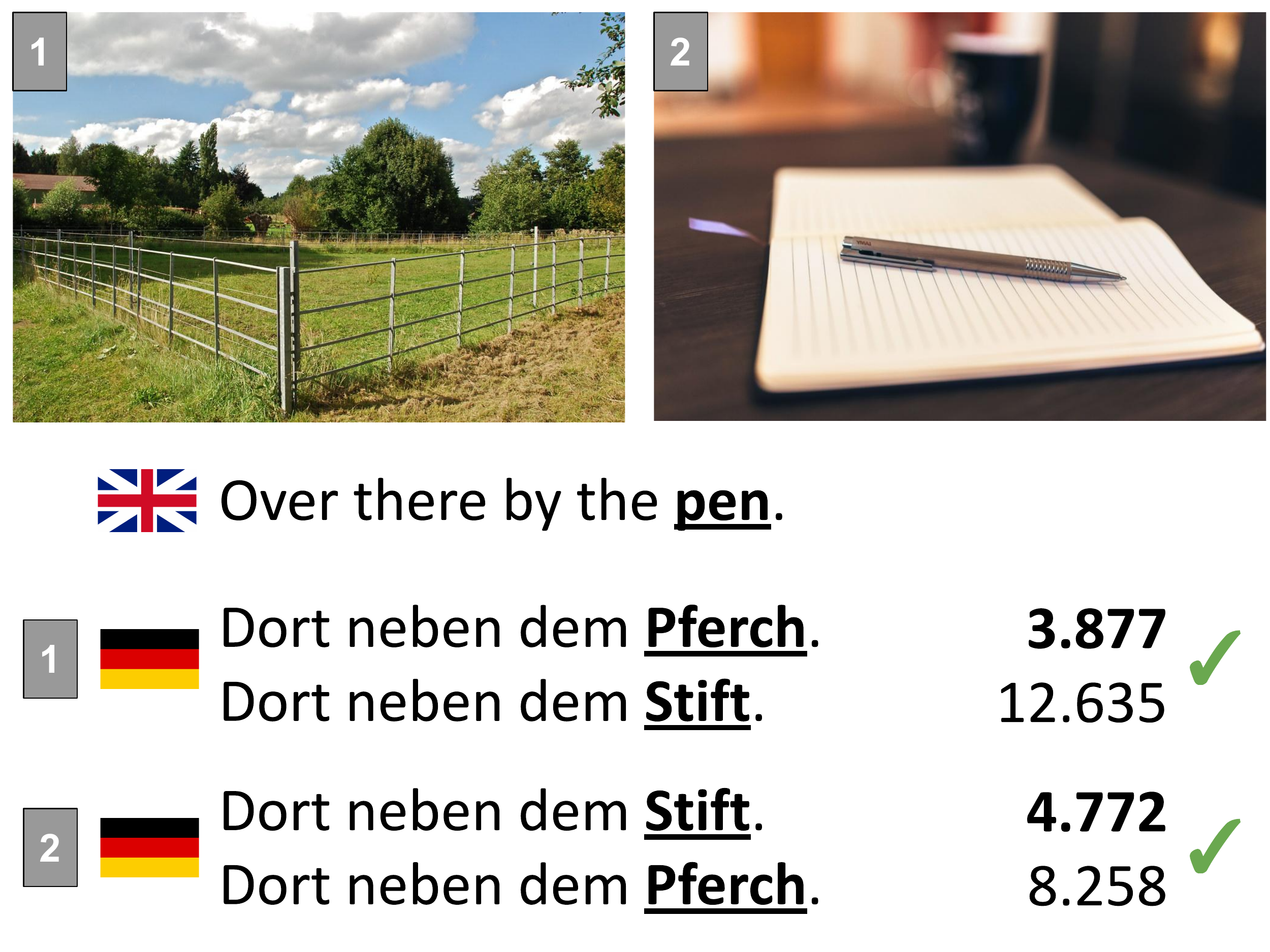}
 \caption{The English word \textit{pen} refers to `an instrument for writing or drawing with ink' or `an area of land surrounded by a fence'.}
    \label{fig:pen_ex}
\end{subfigure}
\hfill
\begin{subfigure}[b]{0.49\textwidth}
  \centering
  \includegraphics[width=.95\linewidth]{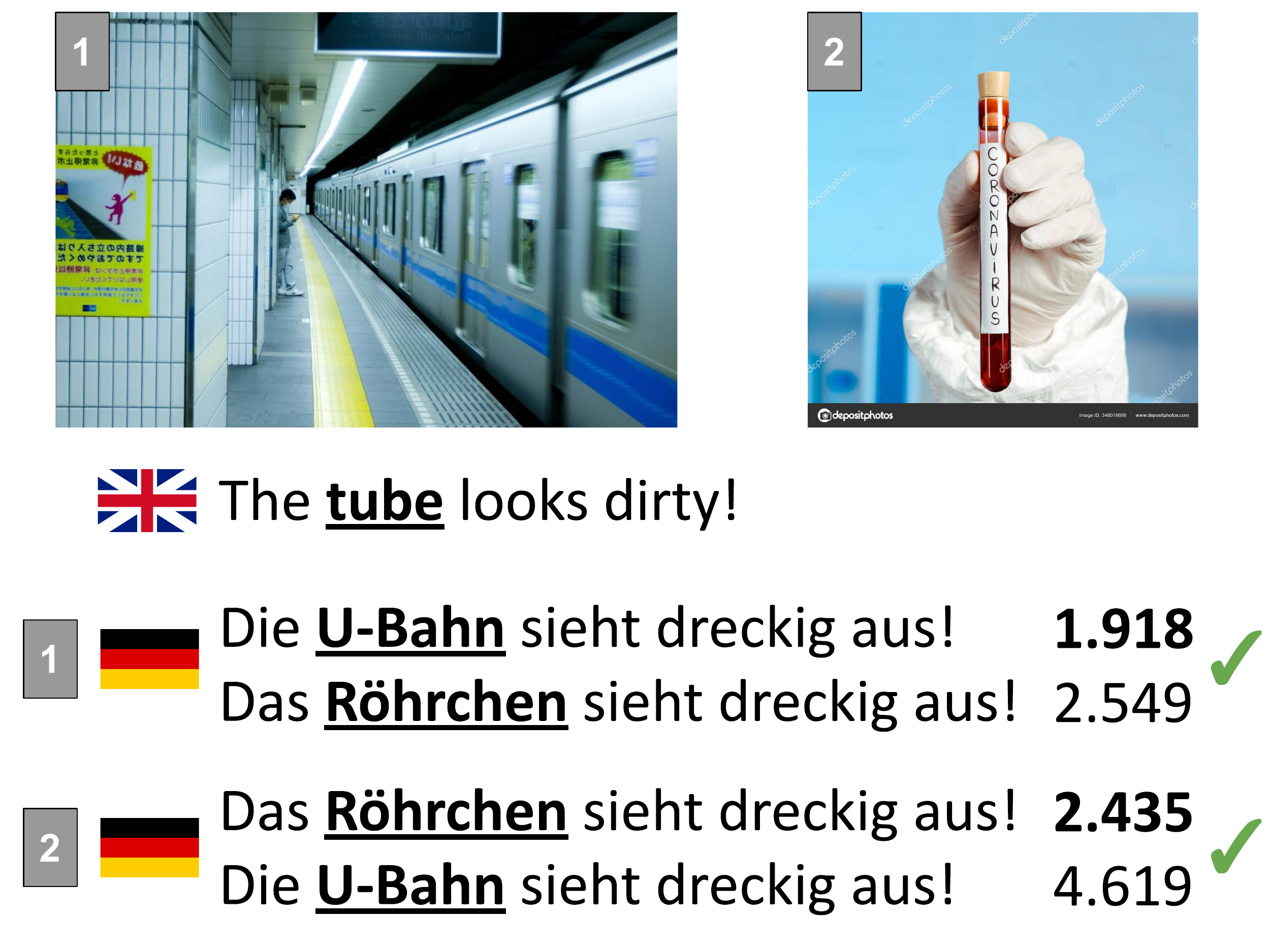}
 \caption{The English word \textit{tube} refers to `a long cylinder for holding liquids or gases.' or `a railway system in some cities'.}
    \label{fig:tube_ex}
\end{subfigure}
\par\bigskip
\caption{Perplexity scores from VGAMT on different examples from CoMMuTE. It is possible to produce at least two different French translations from each source sentence in English, the correct translation therefore depends on the input image. For each sub-example, the correct (resp. incorrect) translation is the top (resp. bottom) one. The ambiguous parts of the sentences are highlighted in bold.}
\label{fig:commute_examples_appendix}
\end{figure*}

Figure~\ref{fig:commute_examples_appendix} shows examples from CoMMuTE and the perplexity scores obtained by VGAMT. It is able to choose the correct translations from English sentences with the ambiguous words \textit{chips}, \textit{bugs}, \textit{red light}. However, it fails to choose the correct translation in the first case of Figure~\ref{fig:beams_ex}; the picture shows a beam `ray of light' and the perplexity of the correct (top) translation with the French translation \textit{rayon} is higher than the incorrect (bottom) one with the French translation \textit{poutre}. Nevertheless, the model gives a lower perplexity to the sentence with the correct image (1.847) in comparison to the same sentence with the incorrect image (2.616). 
So, even if VGAMT is not able to choose the correct translation in the first case of this example, it shows some evidence of being able to discriminate between the French translation with the correct image and the same French translation with the incorrect image. \Cref{fig:pen_ex,fig:tube_ex} show two other similar examples in En$\rightarrow$De MT.

In terms of translation (rather than reranking), Figure~\ref{fig:football_example} shows an example from Multi30k where our model correctly translates the ambiguous word while the text-only baseline fails to do so.
\begin{table}[ht]
\resizebox{\linewidth}{!}{
    \centering\small
    \begin{tabular}{lccc} \toprule
     Model & Test2016 & Test2017 & MSCOCO \\  \midrule

    \multicolumn{4}{c}{\bf Text-only Machine Translation} \\ 
    \midrule

    mBART + MT* \textit{w/ adapters} & 79.9 {\scriptsize±0.3} & 76.0 {\scriptsize±0.2} & \textbf{69.5} {\scriptsize±0.6} \\
    
    \tab \textit{w/o MLM objective} & \textbf{80.3} {\scriptsize±0.2} & \textbf{76.3} {\scriptsize±0.2} & 68.7 {\scriptsize±0.3} \\ \midrule

    \multicolumn{4}{c}{\bf Multimodal Machine Translation} \\ 
    \midrule

    \textbf{VGAMT} (\textit{ours}) & 79.7 {\scriptsize ±0.0}  & 75.9 {\scriptsize ±0.1} & 68.9 {\scriptsize±0.4} \\
        
    \tab \textit{unfrozen w/o adapters} & 79.8 {\scriptsize ±0.5} & 75.8 {\scriptsize ±0.2} & 68.7 {\scriptsize±0.6} \\
    
     \tab \textit{w/o VMLM objective} & 80.3 {\scriptsize ±0.1} & 76.0 {\scriptsize ±0.1} &  68.7 {\scriptsize±0.1} \\ 
     
     \tab \textit{w/o guided self-attention} & 79.6 {\scriptsize ±0.1} & 75.4 {\scriptsize ±0.2} & 68.4 {\scriptsize±0.3} \\ 

     \tab \textit{w/ pretraining (w/o co-training)} & 79.2 {\scriptsize ±0.0} & 74.3 {\scriptsize ±0.1} & 67.9 {\scriptsize±0.2} \\
     
     \tab \textit{w/o MDETR features} & 79.5 {\scriptsize ±0.3} & 75.6 {\scriptsize ±0.1} & 68.9 {\scriptsize±0.6} \\ 
     
      \tab \textit{w/o CLIP features} & 79.2 {\scriptsize±0.5} & 75.2 {\scriptsize ±0.3} & 69.0 {\scriptsize±0.5} \\  \bottomrule
    
    \end{tabular}}
    \caption{METEOR scores for the ablations described in Section~\ref{sec:ablation-study} (En$\rightarrow$Fr). The best result is indicated in \textbf{bold}.}
    \label{tab:meteor-ablation-en-fr}
\end{table}
\section{METEOR scores}\label{app:meteor-scores}

In order to compare to previous work, we also provide METEOR scores in Table~\ref{tab:meteor-scores} for En$\rightarrow$\{Fr,De,Cs\} standard benchmarks. It confirms that VGAMT obtains competitive results over a strong text-only baseline on benchmarks where images are not necessary for translation. METEOR scores for the En$\rightarrow$Fr ablations conducted in Section~\ref{sec:ablation-study} are shown in Table~\ref{tab:meteor-ablation-en-fr}. 

\section{Translating CoMMuTE} \label{app:translating-commute}
\begin{table}[h]
\resizebox{\linewidth}{!}{
    \centering\small
    \begin{tabular}{llcc} \toprule
     &  & VGAMT (\textit{ours}) & mBART + MT* {\small\textit{w/ adapters}}\\ \midrule

    \multirow{3}{*}{En$\rightarrow$Fr} & BLEU & 32.2 {\scriptsize±1.7} & \textbf{34.5} {\scriptsize±1.4} \\
    & COMET & \textbf{0.362} {\scriptsize±0.048} & 0.306 {\scriptsize±0.014} \\
    & METEOR & 48.5 {\scriptsize±2.1} & \textbf{52.3} {\scriptsize±1.4} \\ \midrule
    \multirow{3}{*}{En$\rightarrow$De} & BLEU & \textbf{29.3} {\scriptsize±0.6} & 25.9 {\scriptsize±0.7} \\
    & COMET & \textbf{0.184} {\scriptsize±0.024} & 0.182 {\scriptsize±0.007} \\
    & METEOR & \textbf{43.0} {\scriptsize±0.8} & 41.3 {\scriptsize±1.3} \\ \midrule
    \multirow{3}{*}{En$\rightarrow$Cs} & BLEU & \textbf{20.8} {\scriptsize±0.9} & 18.3 {\scriptsize±1.3} \\
    & COMET & \textbf{0.525} {\scriptsize±0.024} & 0.491 {\scriptsize±0.022} \\
    & METEOR & \textbf{23.4} {\scriptsize±0.8} & 22.4 {\scriptsize±0.7} \\ \bottomrule
    \end{tabular}
    }
    \caption{MT Generation results for CoMMuTE. Best results are indicated in \textbf{bold}.}
    \label{tab:commute-generation} \vspace{-4mm}
\end{table}

\begin{figure}[!h]
    \centering
    \includegraphics[width=\linewidth]{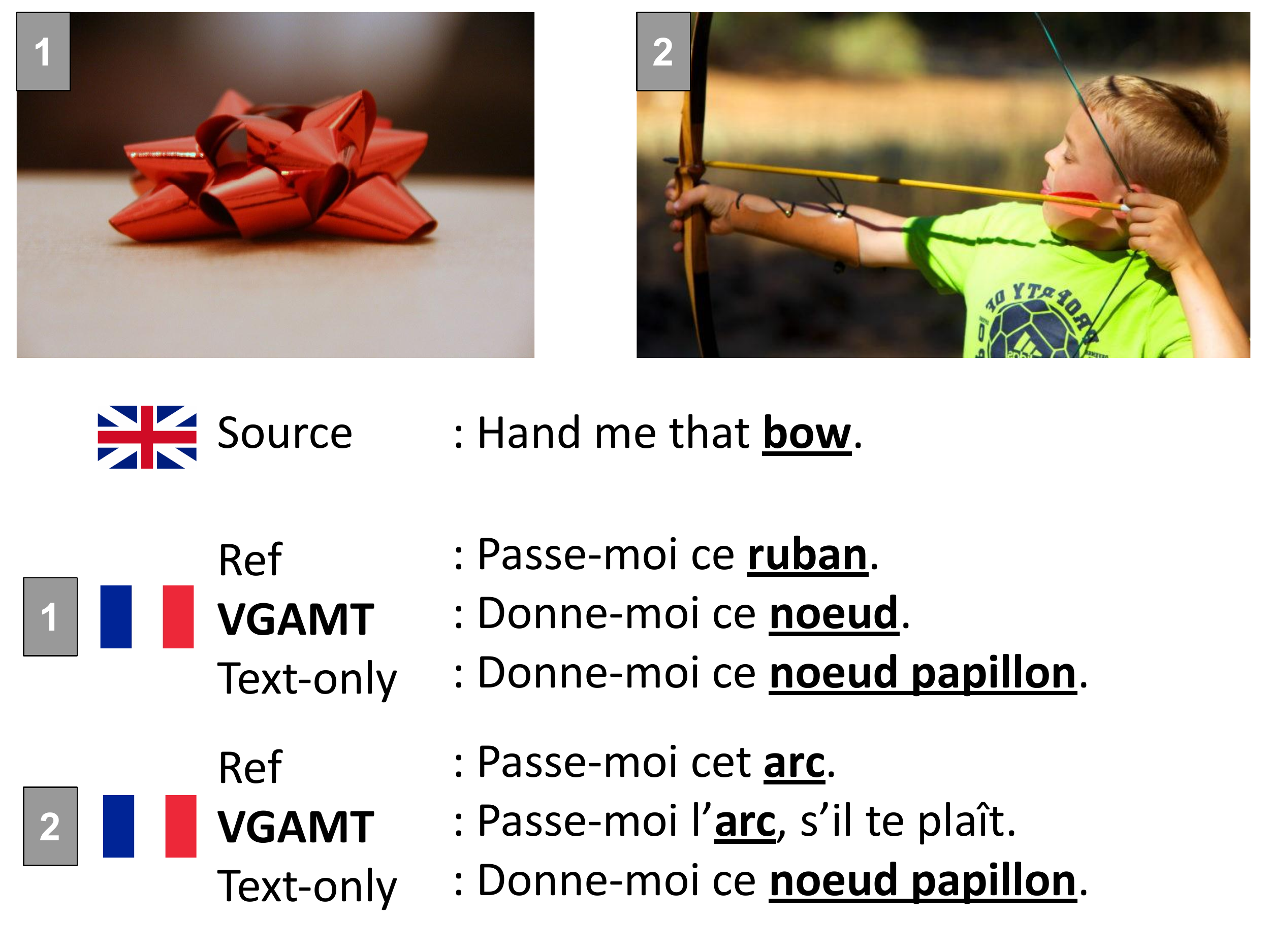}
    \caption{Machine Translation example from CoMMuTE. VGAMT is able to exploit visual information to translate `\textit{bow}' correctly in the two cases.}
    \label{fig:example-bow}
\end{figure}

CoMMuTE is designed as a contrastive test set to be used for reranking. However, it is possible to translate the source sentences too and compare against the reference translations. 

Table~\ref{tab:commute-generation} shows the MT results on CoMMuTE comparing VGAMT and the strong text-only baseline. They may indicate that traditional metrics for MT task are ill-adapted to evaluating the use of visual information by MMT models. For instance, BLEU and METEOR scores for the text-only baseline are significantly higher than the scores for our model VGAMT on the En$\rightarrow$Fr split whereas our VGAMT obtains 67.10 accuracy on the contrastive evaluation (Table~\ref{tab:main_table_fr}). It might be due to the fact that such metrics are less reliable on small datasets or that BLEU and METEOR are words matching metrics and therefore output low scores for synonyms or similar content described differently. On the other hand, COMET is an embedding-based metric, which outputs higher scores for synonyms which may be why VGAMT outperforms the text-only baseline with this metric; as illustrated by Figure~\ref{fig:example-bow} where VGAMT outputs \textit{noeud} `bow' which is a synonym of the reference translation \textit{ruban} `bow' in that case.
That being said, the use of our contrastive dataset CoMMuTE therefore seems necessary to evaluate how well a MMT model exploits visual information in order to produce correct translations instead of relying only on standard metrics for MT.

\begin{figure*}[t]
\centering\small
\begin{subfigure}[b]{0.49\textwidth}
    \centering
    \includegraphics[width=.95\linewidth]{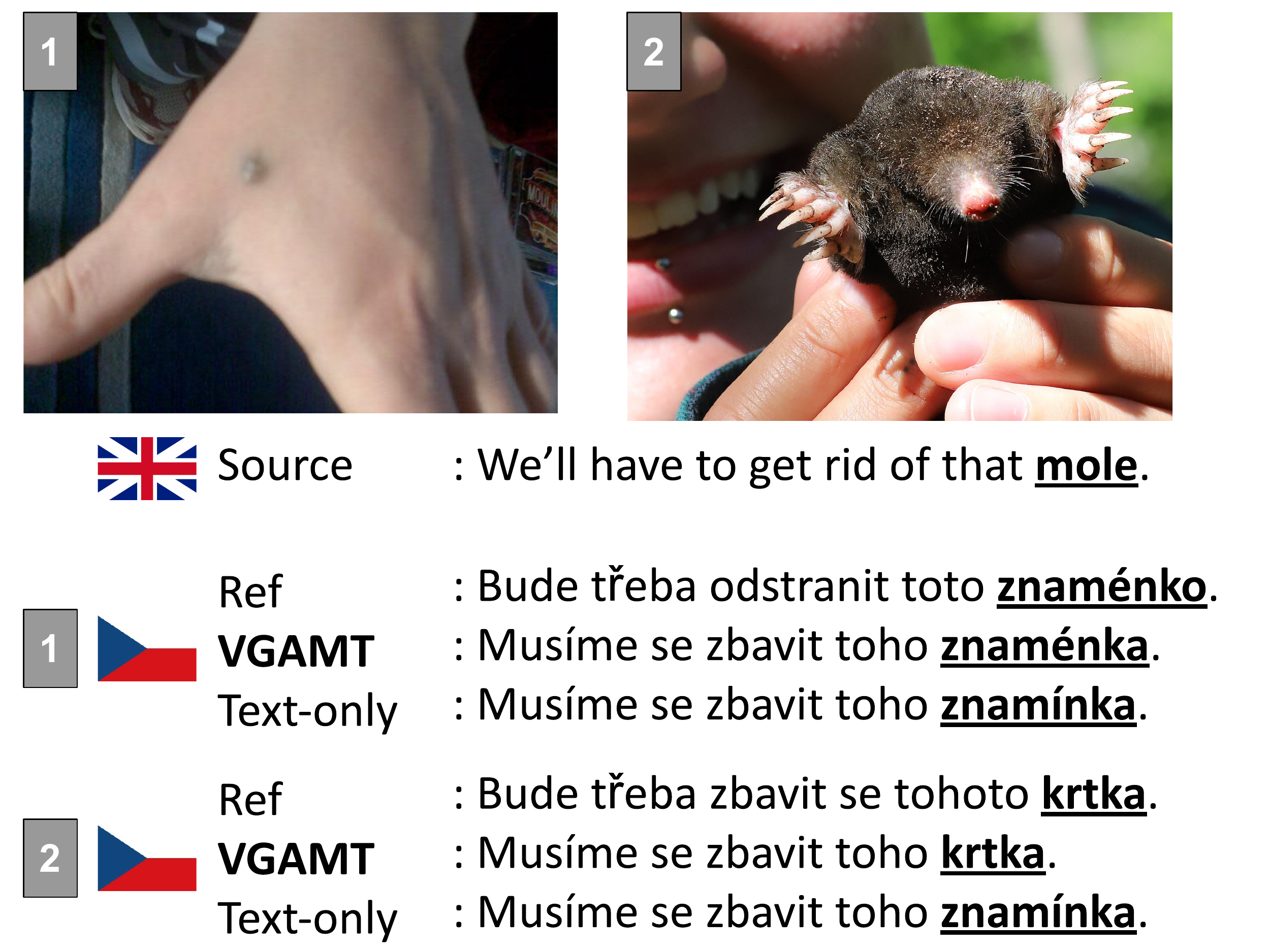}
    \caption{English word \textit{mole} correctly translated in both cases (\textit{znaménko} `skin blemish' and \textit{krtka} `burrowing mammal').}
    \label{fig:example-mole-trad}
\end{subfigure}
\hfill
\begin{subfigure}[b]{0.49\textwidth}
    \centering
    \includegraphics[width=\linewidth]{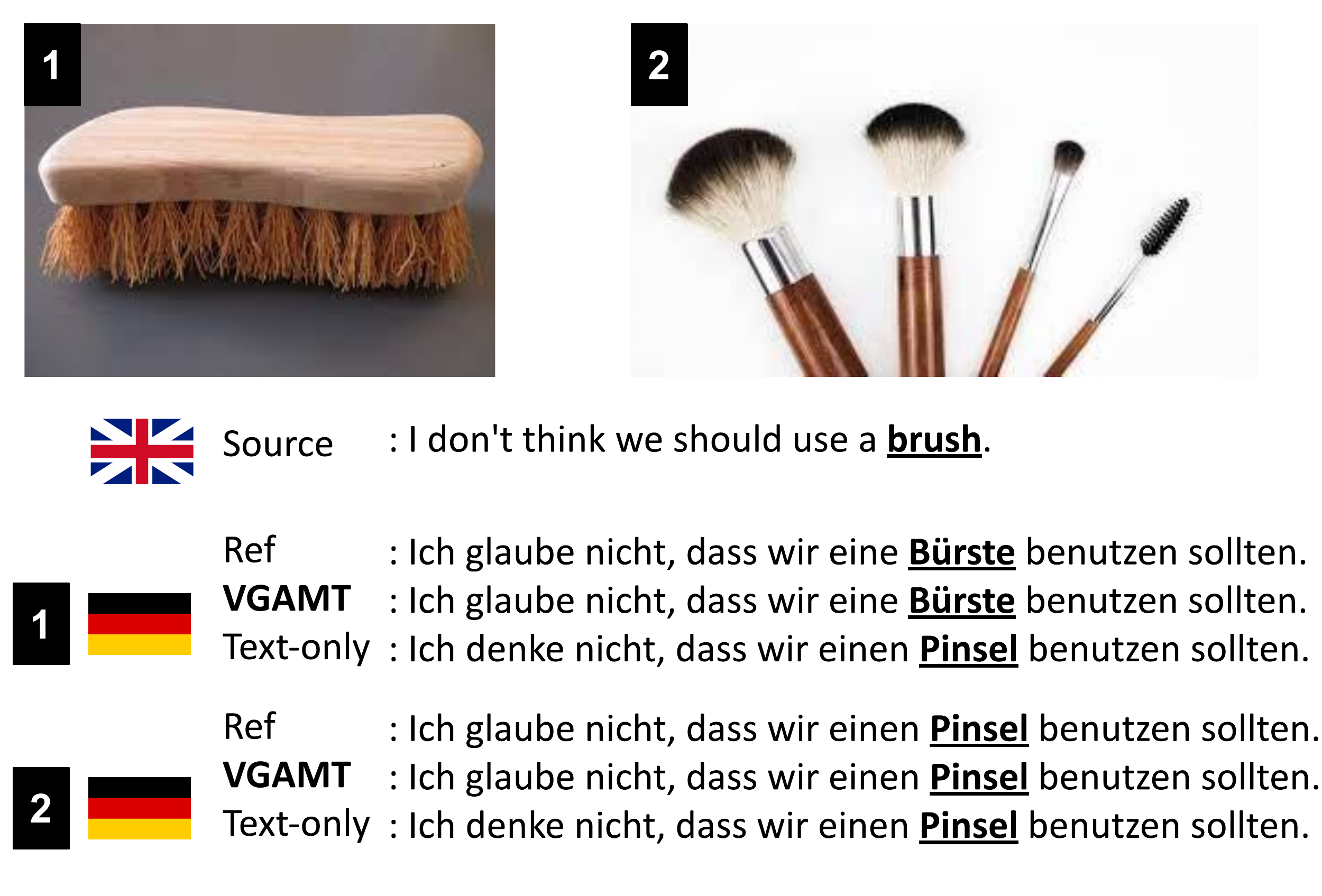}
    \caption{English word \textit{brush} correctly translated in both cases (\textit{Bürste} `cleaning tool' and \textit{Pinsel} `object used for painting').}
    \label{fig:example-arms-trad}
\end{subfigure}
\par\bigskip 
\begin{subfigure}[b]{0.49\textwidth}
    \centering
    \includegraphics[width=.95\linewidth]{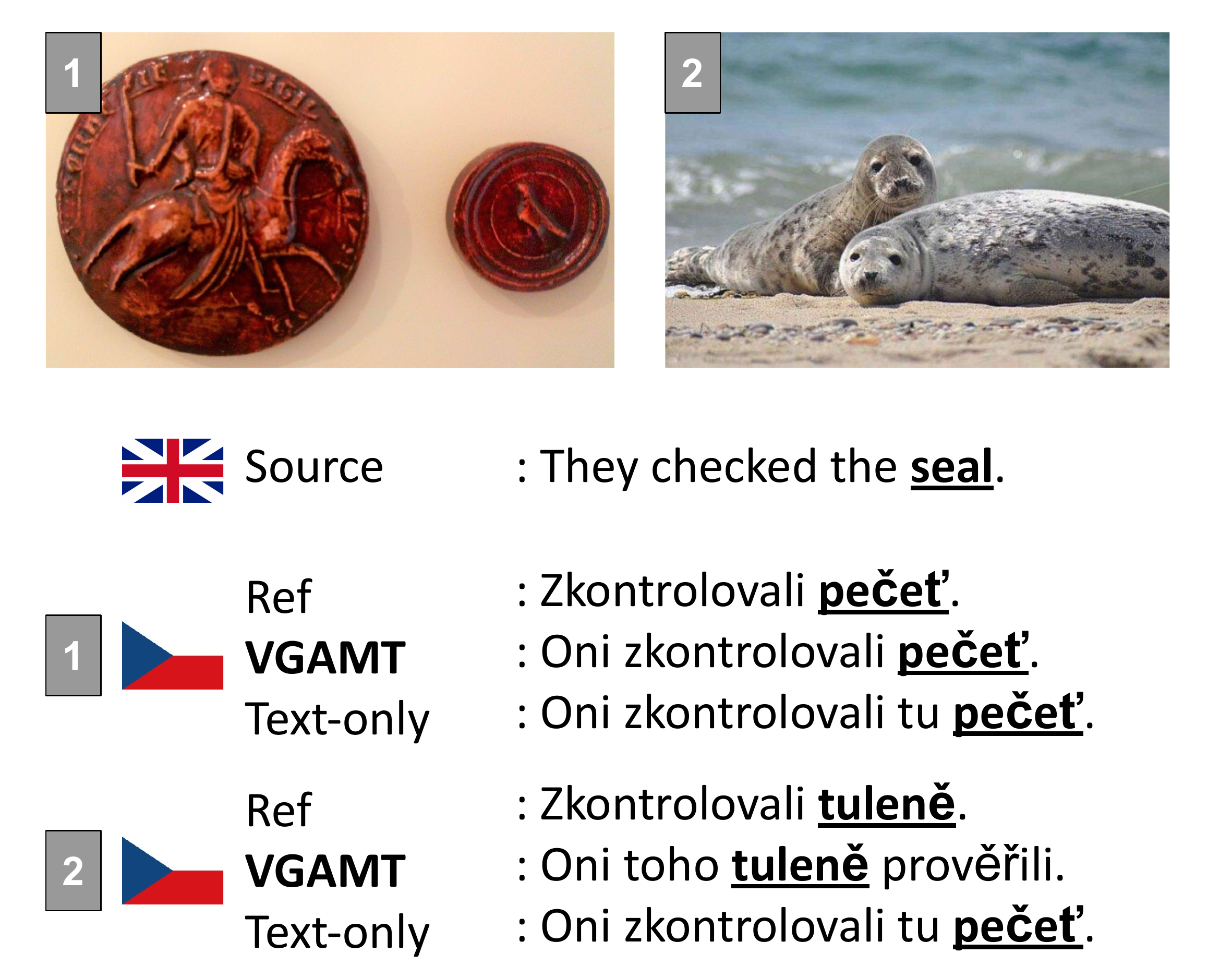}
    \caption{English word \textit{seal} correctly translated in both cases (\textit{pečeť} `official stamp' and \textit{tuleně} `sea mammal').}
    \label{fig:example-seal-trad}
\end{subfigure}
\hfill
\begin{subfigure}[b]{0.49\textwidth}
    \centering
    \includegraphics[width=.95\linewidth]{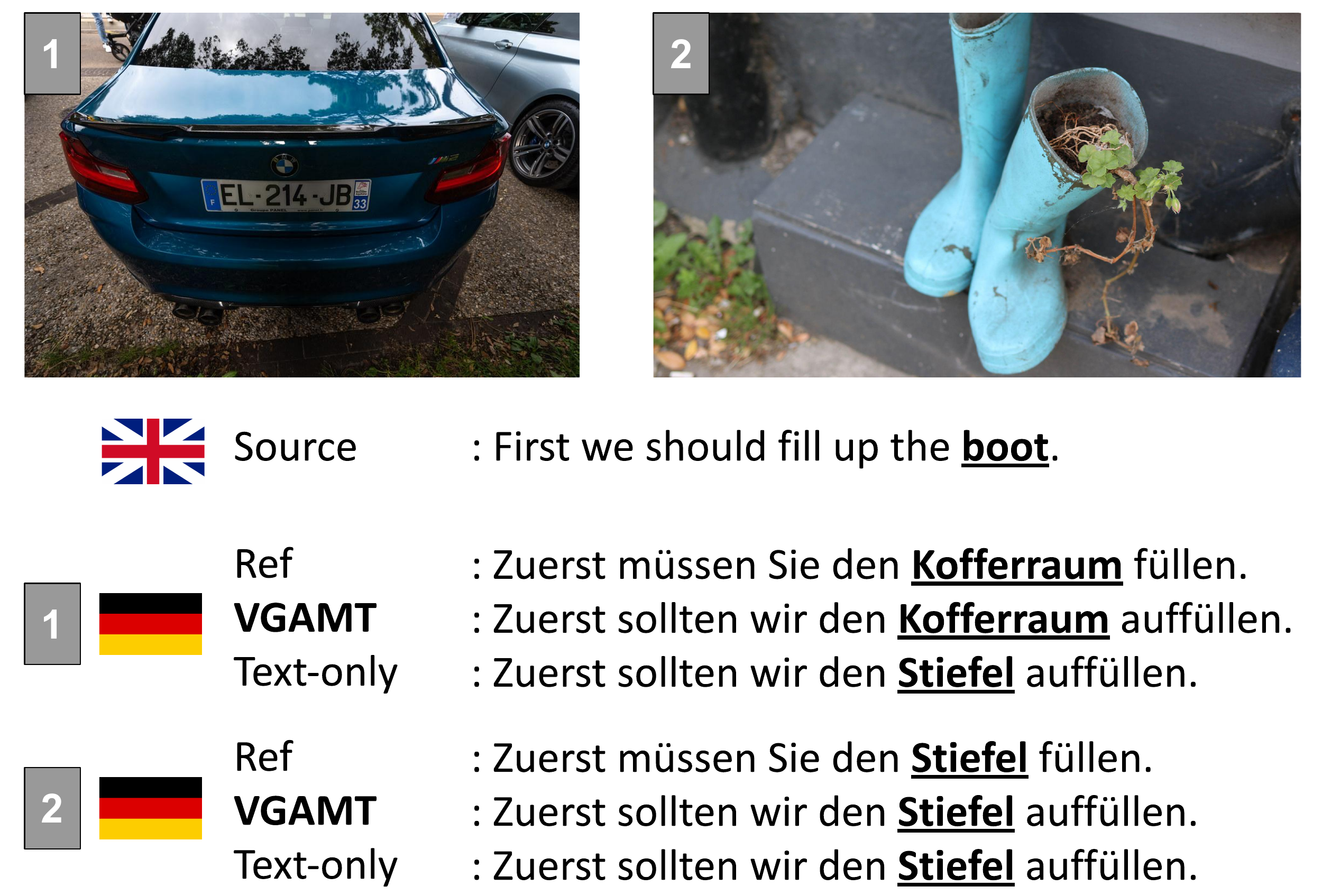}
    \caption{English word \textit{boot} correctly translated in both cases (\textit{Kofferraum} `car trunk' and \textit{Stiefel} `footwear').}
    \label{fig:example-boot-trad}
\end{subfigure}
\par\bigskip 
\begin{subfigure}[b]{0.49\textwidth}
    \centering
    \includegraphics[width=\linewidth]{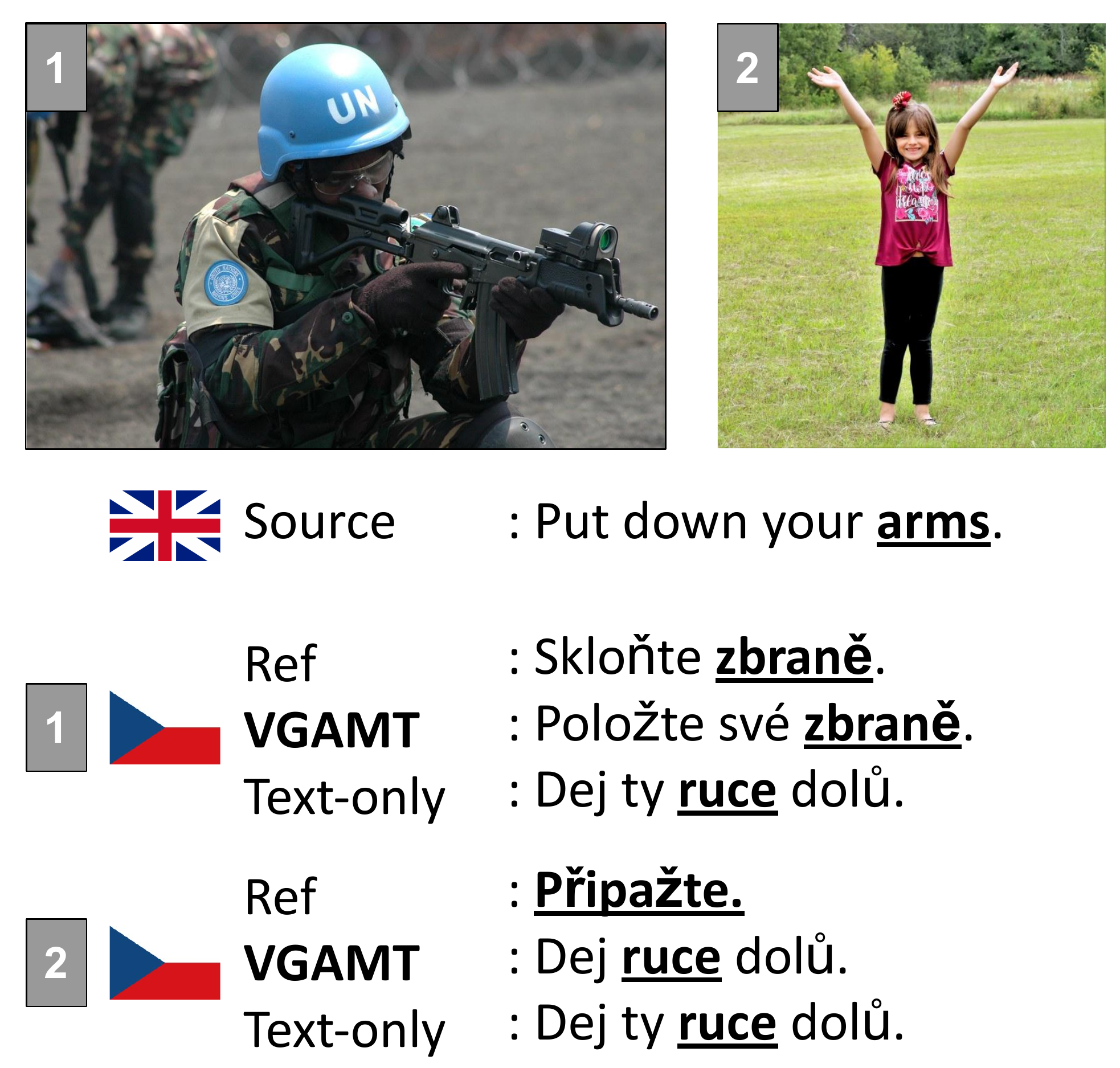}
    \caption{English word \textit{arms} correctly translated in both cases (\textit{zbraně} `weapon' and \textit{ruce} `parts of the human body').}
    \label{fig:example-brush-trad}
\end{subfigure}
\hfill
\begin{subfigure}[b]{0.49\textwidth}
    \centering
    \includegraphics[width=\linewidth]{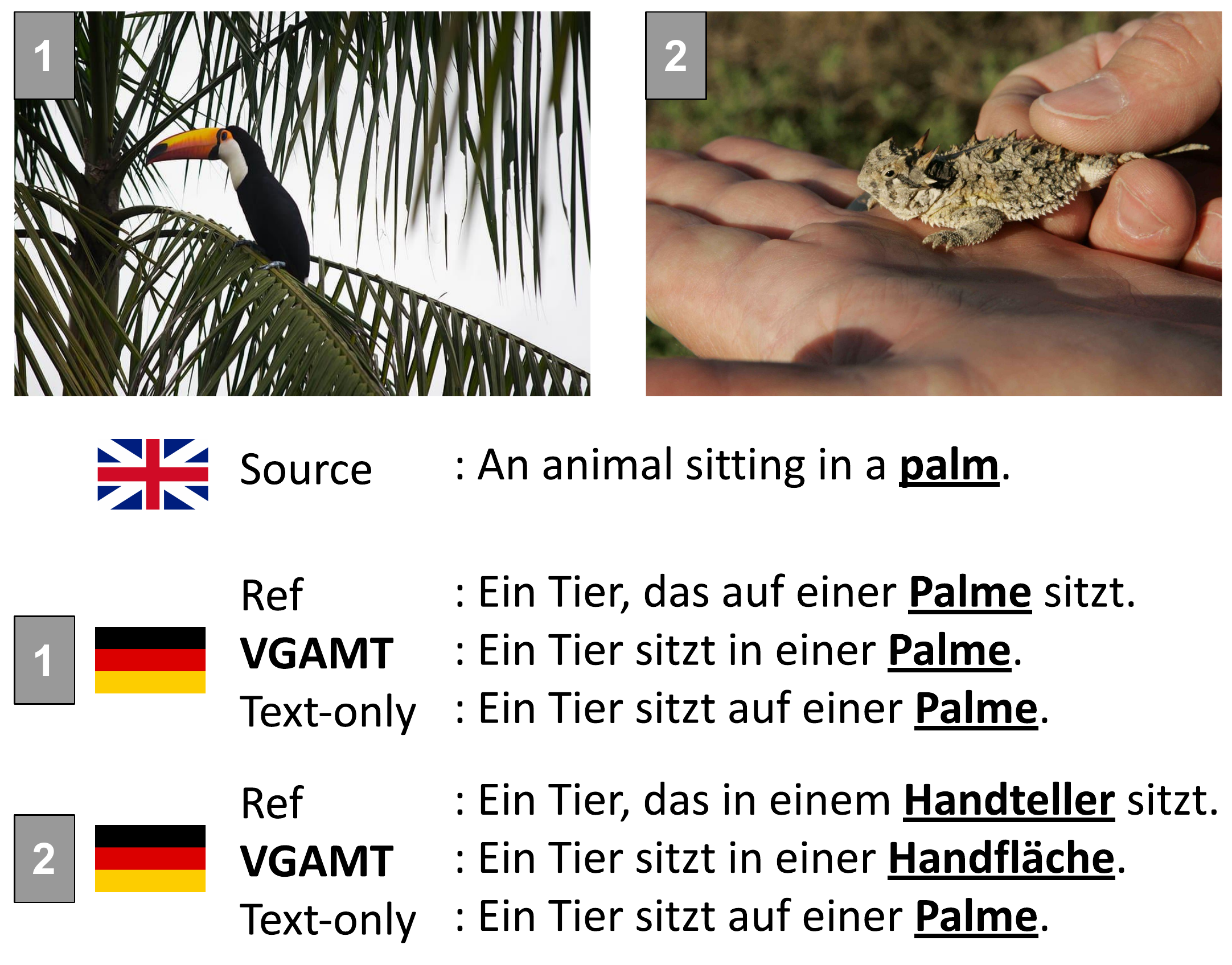}
    \caption{English word \textit{palm} correctly translated in both cases (\textit{Palme} `tree' and \textit{Handfläche} `anterior aspect of the hand').}
    \label{fig:example-palm-trad}
\end{subfigure}
\par\bigskip
\caption{MT examples for different English$\rightarrow$Czech and English$\rightarrow$German examples from CoMMuTE. For each one, VGAMT is able to exploit visual information to translate English ambiguous words (underlined and in bold) correctly in all cases.}
\label{fig:commute_examples_trad_appendix}
\end{figure*}

Figure~\ref{fig:example-bow} illustrates how VGAMT can translate ambiguous words correctly by using images, while mBART + MT (our strong text-only baseline) cannot. In both cases, the baseline outputs French \textit{noeud papillon} `bow tie', while VGAMT produces the correct translations of \textit{bow}. \Cref{fig:example-mole-trad,fig:example-arms-trad,fig:example-seal-trad,fig:example-boot-trad,fig:example-brush-trad,fig:example-palm-trad} show the same effect for En$\rightarrow$\{Cs,De\} translations. Even if VGAMT does not literally translate the ambiguous word as exemplified by Figure~\ref{fig:example-arms-trad}, it produces a translation with the expected meaning based on the image; the text-only models were not able to do so. 

\end{document}